\documentclass{article}

\PassOptionsToPackage{numbers, compress}{natbib}

\usepackage[main,final]{neurips_2026}

\usepackage[utf8]{inputenc} % allow utf-8 input
\usepackage[T1]{fontenc}    % use 8-bit T1 fonts
\usepackage{hyperref}       % hyperlinks
\usepackage{url}            % simple URL typesetting
\usepackage{booktabs}       % professional-quality tables
\usepackage{amsfonts}       % blackboard math symbols
\usepackage{nicefrac}       % compact symbols for 1/2, etc.
\usepackage{microtype}      % microtypography
\usepackage{graphicx}       % figures
\usepackage{subcaption}     % subfigures
\usepackage{multirow}       % multirow cells in tables
\usepackage{wrapfig}        % text-wrapped figures
\PassOptionsToPackage{table}{xcolor}
\usepackage{xcolor}         % colors (with table option for \rowcolor)

\usepackage{amsmath}
\usepackage{amssymb}
\usepackage{mathtools}
\usepackage{amsthm}
\usepackage[most]{tcolorbox}
\usepackage{fvextra}
\DeclareMathOperator*{\argmax}{arg\,max}
\newcommand{\wltsrow}{\rowcolor{black!5}}
\usepackage{algorithm}
\usepackage{algorithmic}

\theoremstyle{plain}
\newtheorem{theorem}{Theorem}[section]

\newtheorem{lemma}[theorem]{Lemma}
\newtheorem{corollary}[theorem]{Corollary}
\theoremstyle{definition}

\theoremstyle{remark}

\title{Backtracking with Linear-in-Depth \\ Search-Space Growth:
Width-Limited Tree Search \\ for Large Language Models}

\author{%
  Jiacheng Xu\textsuperscript{1,2} \quad
  Fuxiang Zhang\textsuperscript{1,2} \quad
  Chaojie Wang\textsuperscript{2} \quad
  Bo An\textsuperscript{1} \\
  \textsuperscript{1}Nanyang Technological University, Singapore \qquad
  \textsuperscript{2}Skywork AI \\
  \texttt{jiacheng005@e.ntu.edu.sg, boan@ntu.edu.sg}
}

\begin{document}

\maketitle

\begin{abstract}
Inference-time search offers a principled way to improve the reliability of large language models (LLMs) by exploring and scoring intermediate trajectories.
However, widely used search paradigms exhibit complementary failure modes: sequence search scales linearly but commits early and cannot recover from mistaken prefixes, whereas tree search can backtrack but often becomes impractical for deep search due to the exponential growth of the search space with depth.
These limitations become increasingly pronounced in long-chain-of-thought responses.
We propose \emph{Width-Limited Tree Search} (WLTS), which reconciles these strengths by preserving the tree-search structure that enables backtracking while using a dynamic expandability criterion to enforce a strict per-layer width constraint.
This design keeps generated search-tree growth linear in depth and sustains verifier supervision across the full reasoning process.
WLTS further supports two optional deployment components: prefix deletion for diversity and agreement-based early stopping to reduce computation on tasks with canonicalizable outputs.
Across mathematical, coding, and logical reasoning benchmarks, and under both classifier and generative verifier settings, WLTS achieves higher accuracy and better inference-budget efficiency than well-established sequence- and tree-search baselines.
\end{abstract}

\section{Introduction}

\begin{wrapfigure}[17]{r}{0.5\textwidth}
   \vspace{-2.5em}
   \centering
   \includegraphics[width=0.45\textwidth]{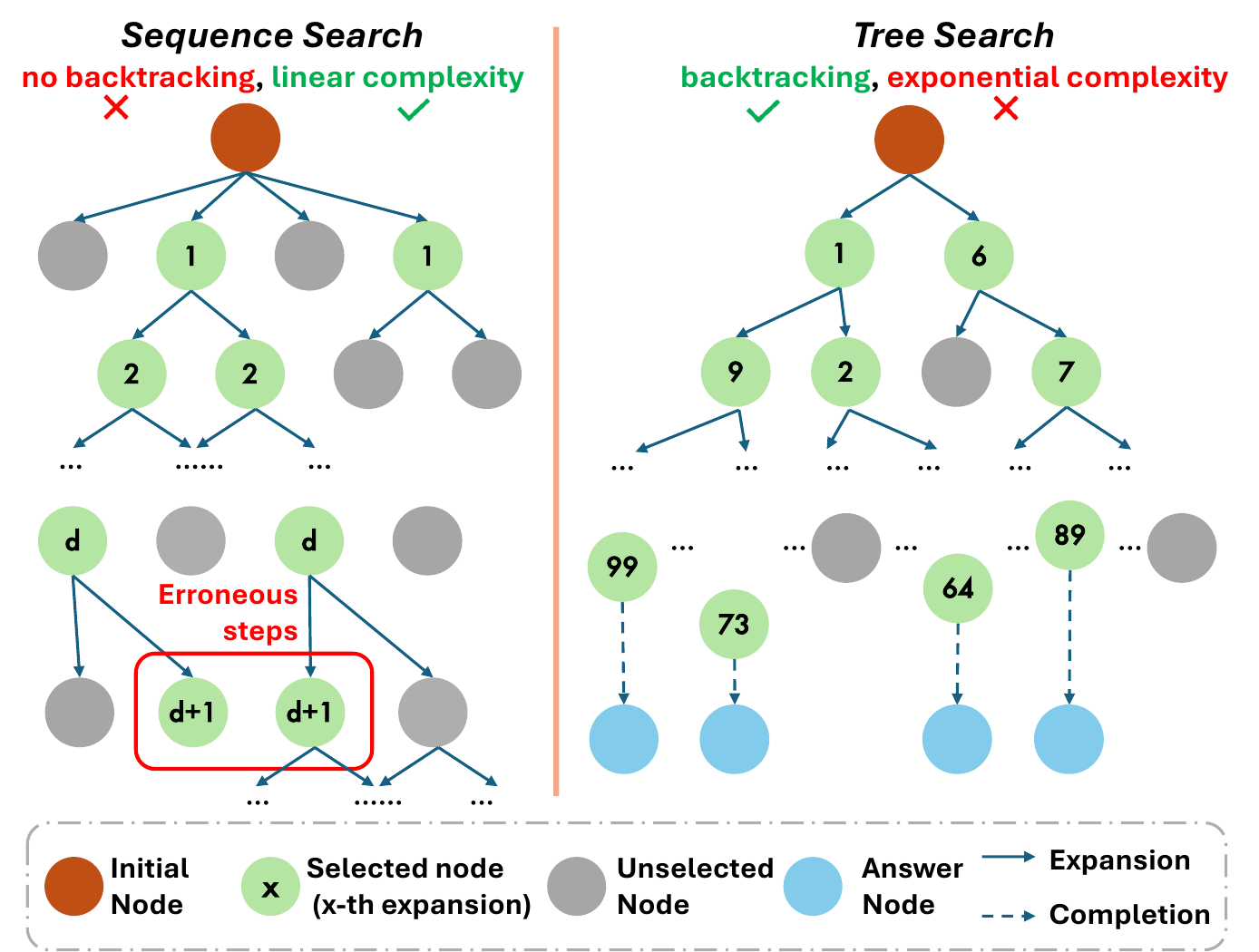}
   \caption{Limitations of sequence and tree search. The left may get stuck in erroneous intermediate nodes and cannot backtrack. The right experiences exponential search-space growth with depth, so limited resources restrict it to shallow searches.}
   \label{fig:motivation}
   \vspace{-2em}
\end{wrapfigure}

Large language models (LLMs) have demonstrated remarkable capabilities and have been widely adopted across diverse applications~\citep{openai2023gpt4,DeepSeekR12025}.
However, the quality of responses generated in a single-pass, serial manner is often unstable, which has led to increasing attention on techniques for improving inference-time performance, especially for multi-step reasoning tasks~\citep{wei2022cot,OptimalScaling2024,ERPO2026}.
Among these, search algorithms stand out by simultaneously exploring multiple candidate intermediate trajectories and selecting the most promising ones using a reward model or verifier~\citep{lightman2023verify,Heimdall2025,GenerativeVerifiers2025,TCS2026}.

From the perspective of backtracking, search algorithms for LLMs can be broadly divided into sequence search and tree search~\citep{SearchSurvey2025,TreeSearchSurvey2025}.
As shown in Figure~\ref{fig:motivation}, sequence search, such as best-of-N and beam search~\citep{OptimalScaling2024}, does not allow backtracking: once the candidate set is saturated with erroneous steps, the model can only continue reasoning from those mistakes.
In contrast, tree search, including Monte Carlo Tree Search (MCTS)~\citep{browne2012mcts} and recent LLM-oriented variants~\citep{yao2023tot,rStarMath2025}, explores multiple continuations from each intermediate state and can return to an earlier decision point when later evidence reveals an unpromising branch.
This backtracking improves fault tolerance by allowing the search to abandon erroneous trajectories and try alternative continuations.
However, without structural control, the number of generated search-tree nodes can grow as $O(b^d)$ with branching factor $b$ and reasoning depth $d$.
Under limited inference budgets, this exponential growth often causes tree search to over-explore shallow layers and rely on single-pass completion for the remaining reasoning steps~\citep{Qstar2024,WiderDeeper2025}.
This phenomenon becomes more pronounced as reasoning horizons, model sizes, and context lengths increase~\citep{DonGetLost2025}.

Because backtracking offers stronger fault tolerance and a higher performance ceiling, much recent work on LLM inference has focused on tree-search methods that allocate inference-time computation more effectively~\citep{ETS2025}.
Existing approaches are often guided by two signals: \emph{value-guided} allocation uses reward models or verifiers to score partial trajectories and prune nodes or prioritize expansions~\citep{liteSearch2024,Qstar2024,DonGetLost2025}, while \emph{uncertainty-aware} allocation uses uncertainty, variance, or expected information gain to balance exploration and exploitation~\citep{UncertaintyGuidedOptimization2024,UncertaintyOfThoughts2024}.
These directions are often combined, but they mainly reallocate expansion budgets without inherently enforcing a strict per-layer capacity limit.
Consequently, generated search-tree nodes can still grow exponentially with depth, concentrating finite budgets in shallow layers and weakening verifier supervision over deeper reasoning steps~\citep{ETS2025,WiderDeeper2025,DonGetLost2025}.

To address this structural bottleneck, we propose \emph{Width-Limited Tree Search} (WLTS), a tree-search algorithm that preserves backtracking while constraining generated search-tree nodes to grow linearly with reasoning depth.
WLTS retains the standard selection--expansion--evaluation--backpropagation loop, but replaces unconstrained branching with a strict per-layer width cap $W$ enforced by a dynamic expandability criterion.
This criterion permits expansion only {at or beyond the deepest saturated layer}, forcing search budget to propagate along depth rather than accumulate in shallow branches.
Among eligible nodes, WLTS expands the highest-priority node by one reasoning step, sustaining verifier supervision over deeper reasoning while retaining the ability to revise earlier decisions through backtracking.
WLTS also includes two optional deployment components: prefix deletion to reuse freed width for diversity, and agreement-based early stopping to reduce computation on tasks with canonicalizable outputs.

We summarize our contributions as follows:
{\setlength{\leftmargini}{2em}
\begin{itemize}
    \item To the best of our knowledge, WLTS is the first tree-search-style method for LLM inference that supports backtracking while guaranteeing linear-in-depth growth of generated search-tree nodes.
    \item We provide two configurable mechanisms that are naturally compatible with WLTS, supporting practical tradeoffs between accuracy and efficiency.
    \item Through extensive experiments on challenging mathematical, coding, and logical reasoning benchmarks, under both classifier and generative reward model settings, we show that WLTS achieves higher accuracy and better inference-budget efficiency than well-established baselines.
\end{itemize}
}

\begin{figure*}[t]
   \centering
   \includegraphics[width=0.9\textwidth]{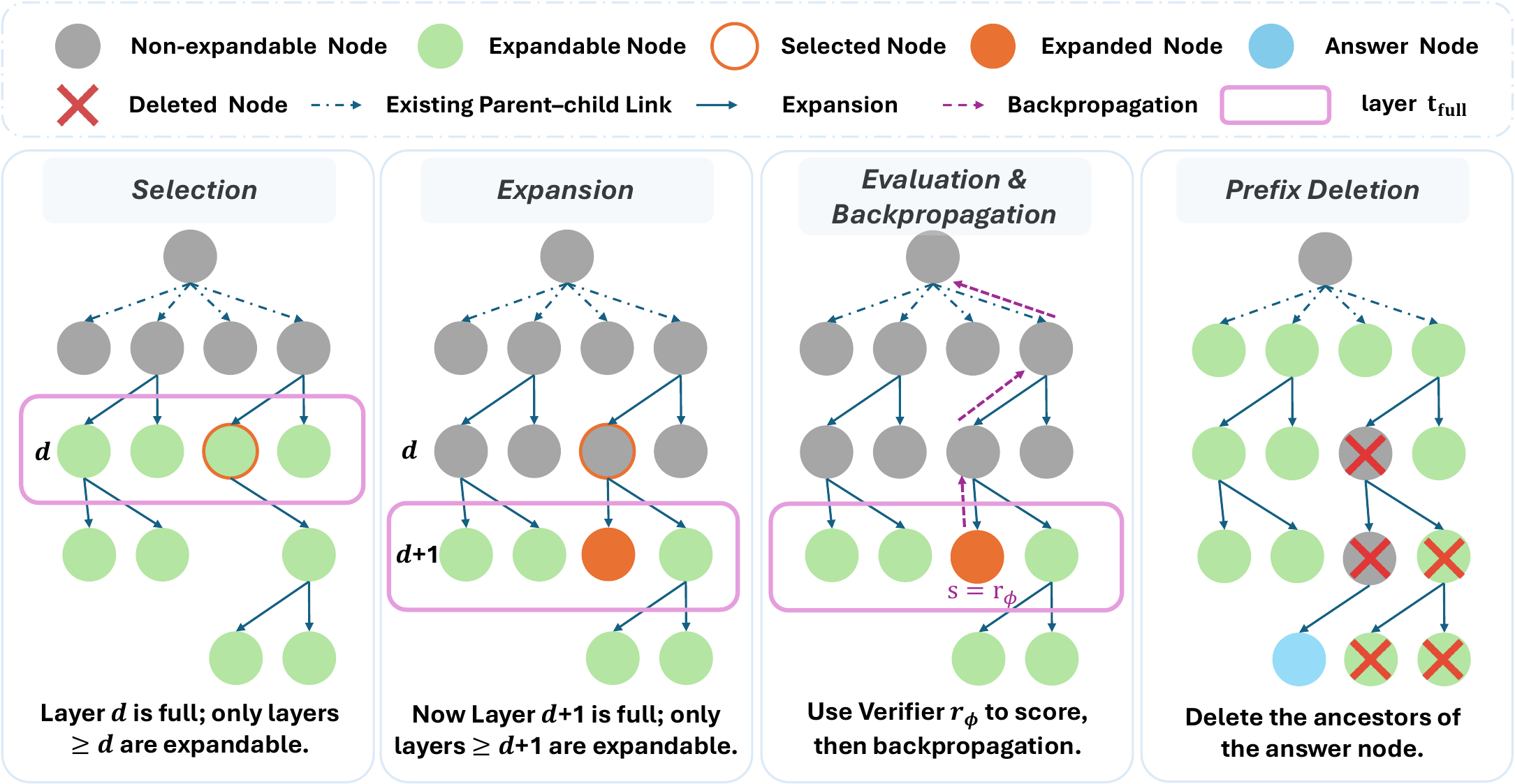}
   \caption{Overview of Width-Limited Tree Search (WLTS). WLTS maintains layer-wise search nodes under a strict per-layer width cap, repeatedly selects an eligible node by priority score, expands it by one reasoning step, evaluates it with a verifier, and backpropagates the score. When a terminal trajectory is found, WLTS may apply optional deletion to free width capacity and diversify exploration.\looseness=-1}
   \label{fig:workflow}
\end{figure*}

\section{Background}

We briefly review the reward models and search algorithms used as points of comparison for WLTS.
Let $x$ denote the question prompt and $y=(y^{(1)},\dots,y^{(d)})$ 
the trajectory segmented into $d$ \emph{steps}, each $y^{(t)}$ being 
an LLM-generated text span, delimited or length-limited.
A partial trajectory $y^{(1:t)}$ defines a node at depth $t$, with the proposer $p_\theta(y^{(t+1)}\mid x,y^{(1:t)})$ generating next-step continuations and the verifier $r_\phi(x,y^{(1:t)})$ scoring partial trajectories to guide search.

\subsection{Reward Model}
\label{sec:grm}
Search algorithms rely on a scoring function to compare partial trajectories and select final outputs~\citep{RewardGuidedTreeSearch2024}.
Given a prompt $x$ and partial trajectory $y^{(1:t)}$, a reward model provides a scalar signal $r_\phi(x,y^{(1:t)})$, either from a trained classifier~\citep{SkyworkReward2024,MathShepherd2024} or from a prompted generative verifier~\citep{RLContemplation2024,GenerativeVerifiers2025,Heimdall2025,SelfVerify2025}.
For the latter, we prompt a verifier LLM to emit a discrete judgment, e.g., \textit{Is the existing reasoning correct (Yes/No)?}, and convert the returned token probability into a scalar score.
Let $\hat{z}\in\{\text{\texttt{Yes}},\text{\texttt{No}}\}$ denote the generated judgment token and let $\ell_\phi(\hat{z}\mid x,y^{(1:t)})$ be its log probability; we use the signed mapping\looseness=-1
\[
r_\phi(x,y^{(1:t)}) =
\begin{cases}
+\exp(\ell_\phi(\hat{z}\mid x,y^{(1:t)})), & \hat{z}=\text{\texttt{Yes}},\\
-\exp(\ell_\phi(\hat{z}\mid x,y^{(1:t)})), & \hat{z}=\text{\texttt{No}}.
\end{cases}
\]
By varying prompt granularity, the same generative reward model can act as an outcome verifier or as a process/progress verifier that provides dense per-step feedback~\citep{lightman2023verify}.

\subsection{Search Algorithms}

\paragraph{Beam Search.}
Beam search is a standard width-limited \emph{sequence} search method, commonly used as a baseline for LLM inference~\citep{sutskever2014seq2seq}.
It maintains a beam $\mathcal{B}_t$ of up to $W$ partial trajectories at depth $t$, expands each by $b$ next-step continuations, scores the resulting candidates, and keeps the top $W$ as $\mathcal{B}_{t+1}$.
This yields a predictable $O(Wbd)$ candidate evaluation cost at depth $d$, but pruning is irreversible: with noisy step-level scores or correlated proposals, the beam can collapse onto an incorrect path and force subsequent search to extend it.
WLTS inherits the benefit of explicit width control, but keeps a tree structure so that earlier prefixes can still be revisited through backtracking.

\paragraph{Monte Carlo Tree Search.}
Monte Carlo Tree Search (MCTS) incrementally grows a search tree by repeatedly running simulations and reallocating computation toward promising branches~\citep{browne2012mcts,Qstar2024}.
In the LLM setting, nodes are partial trajectories and edges are step-level continuations; each simulation follows selection, expansion, evaluation, and backpropagation.
A common selection rule is UCT~\citep{kocsis2006uct}: given a node $y^{(1:t)}$ with expanded children $\mathrm{C}(y^{(1:t)})$, visit count $n(y^{(1:t)})$, and empirical mean return $v(y^{(1:t)})$, UCT selects the next child by
\[
y^{(1:t+1)\,*} = \argmax_{y'\in \mathrm{C}(y^{(1:t)})}
\left(
v(y') + c_{\mathrm{uct}} \sqrt{\frac{\log n(y^{(1:t)})}{n(y')+1}}
\right),
\]
where $c_{\mathrm{uct}}>0$ controls exploration.

Unlike beam search, MCTS enables \emph{backtracking}: as evidence accumulates, it can revisit earlier prefixes and reallocate budget to alternative branches.
However, without structural constraints, generated search-tree nodes can scale as $O(b^d)$ with branching factor $b$ and depth $d$.
Under finite budgets, this growth often confines exploration to shallow layers.
WLTS retains this backtracking capability while enforcing per-layer width control to keep generated search-tree growth linear in depth. \looseness=-1

\section{Width-Limited Tree Search}
\label{sec:wlts}

\subsection{Motivation}
We aim to retain the ability to backtrack, which is the key advantage of tree search for LLM reasoning, without allowing the generated search tree to grow exponentially with depth.
Figure~\ref{fig:workflow} provides an overview of WLTS.
The central design principle is to enforce a strict \emph{per-layer} width cap, forcing search resources to propagate along depth rather than accumulate in shallow branches.
Intuitively, WLTS can revisit and extend earlier reasoning states whenever later evidence suggests that a branch is unpromising, as long as it satisfies the width constraint.

\subsection{Search Procedure}
Each node in the search tree corresponds to a partial trajectory $y^{(1:t)}$.
For each depth $t$, let $\mathcal{L}_t$ collect the nodes kept at that depth.
WLTS enforces the strict per-layer width constraint
\[
|\mathcal{L}_t|\le W,\quad \forall t,
\]
For each node $y^{(1:t)}$, WLTS stores a visit count $n(y^{(1:t)})$ and a value estimate $v(y^{(1:t)})$, the empirical mean of verifier rewards observed from expansions through this node.
The search repeats selection--expansion--evaluation--backpropagation until it collects $N$ terminal trajectories or meets a stopping criterion, and returns the best terminal trajectory by verifier score.

\paragraph{Initialization.}
The root node is the empty trajectory $\varepsilon$ (depth $0$), whose associated state includes the prompt $x$.
Initially, $\mathcal{L}_0=\{\varepsilon\}$ and $\mathcal{L}_t=\emptyset$ for all $t\ge 1$.
We initialize $n(\varepsilon)=1$ and $v(\varepsilon)=0$ so that the UCT bonus $\log n(\varepsilon)$ is well-defined.
The search hyperparameters are the width cap $W$, depth limit $D_{\max}$, expansion budget $B_{\max}$, terminal-candidate budget $N$, and deletion depth $h_{\mathrm{del}}$ used by the optional deletion mechanism in Section~\ref{sec:wlts:diversity}.

\paragraph{Selection.}
WLTS uses a dynamic \emph{expandability} rule to enforce the per-layer width cap.
We call a layer $t\ge 1$ \emph{saturated} if $|\mathcal{L}_t|=W$, and define $t_{\mathrm{full}}$ as the deepest saturated layer; if no layer is saturated, we set $t_{\mathrm{full}}=0$:
\[
t_{\mathrm{full}} \triangleq \max\Bigl(\{t\ge 1:\ |\mathcal{L}_t|=W\}\cup\{0\}\Bigr).
\]
Expanding a depth-$t$ node creates exactly one child at depth $t+1$.
By definition of $t_{\mathrm{full}}$, every layer strictly deeper than $t_{\mathrm{full}}$ is \emph{not} saturated.
Therefore, restricting expansion to depths $t\ge t_{\mathrm{full}}$ guarantees that any child inserted into $\mathcal{L}_{t+1}$ lands in a layer with remaining width capacity.
We define the expandable set as
\[
\mathcal{E} \triangleq \bigcup_{t\ge t_{\mathrm{full}}}\mathcal{L}_t.
\]
Among expandable nodes, WLTS selects using an exploration--exploitation score,
\[
y^\star \,=\, \argmax_{y\in\mathcal{E}}
\left(
v(y) + c_{\mathrm{uct}} \sqrt{\frac{\log n(\varepsilon)}{n(y)+1}}
\right),
\]
where $c_{\mathrm{uct}}\ge 0$ controls exploration and $n(\varepsilon)$ is the root visit count, which tracks the total number of expansions so far.
Setting $c_{\mathrm{uct}}=0$ recovers greedy best-first selection; ties can be broken by preferring deeper nodes.

\paragraph{Expansion.}
Given the selected node $y^{(1:t)}$, WLTS samples a next-step continuation from the proposer,
$y^{(t+1)}\sim p_\theta(\cdot\mid x,y^{(1:t)})$, and forms the child node $y^{(1:t+1)}$.
In our implementation, a step is terminated either by a designated step delimiter or by a fixed token budget.

\paragraph{Evaluation and Backpropagation.}
The new node is scored by the verifier reward model,
$s \triangleq r_\phi(x,y^{(1:t+1)})$ (Section~\ref{sec:grm}).
We set $n(y^{(1:t+1)})\leftarrow 1$ and $v(y^{(1:t+1)})\leftarrow s$.
The score $s$ is then propagated upward along the path to the root.
For each ancestor node $u$ on this path, we update its visit count and running-average value estimate by
\[
n(u)\leftarrow n(u)+1,\qquad
v(u)\leftarrow v(u) + \frac{s-v(u)}{n(u)}.
\]
If the child trajectory is terminal or reaches $D_{\max}$, it is added to the terminal candidate set $\mathcal{T}$.
Otherwise, the child node is inserted into the corresponding layer $\mathcal{L}_{t+1}$.
Finding a terminal trajectory may also trigger optional deletion as described below.

\begin{algorithm}[t]
\caption{Width-Limited Tree Search (WLTS)}
\label{alg:wlts}
\begin{algorithmic}[1]
\REQUIRE prompt $x$, proposer $p_\theta$, verifier $r_\phi$, width cap $W$, depth limit $D_{\max}$, terminal target $N$, expansion budget $B_{\max}$, exploration coefficient $c_{\mathrm{uct}}$, deletion depth $h_{\mathrm{del}}$
\STATE Initialize $\mathcal{L}_0\leftarrow\{\varepsilon\}$, $\mathcal{L}_t\leftarrow\emptyset$ for $t\ge 1$, and $\mathcal{T}\leftarrow\emptyset$
\STATE Initialize $n(\varepsilon)\leftarrow 1$ and $v(\varepsilon)\leftarrow 0$
\FOR{$k=1$ to $B_{\max}$}
\STATE $t_{\mathrm{full}}\leftarrow \max\left(\{t\ge 1:|\mathcal{L}_t|=W\}\cup\{0\}\right)$
\STATE $\mathcal{E}\leftarrow \bigcup_{t\ge t_{\mathrm{full}}}\mathcal{L}_t$
\STATE Select $y^{(1:t)}\leftarrow \argmax\limits_{y\in\mathcal{E}} \left(v(y) + c_{\mathrm{uct}} \sqrt{\frac{\log n(\varepsilon)}{n(y)+1}}\right)$
\STATE Sample $y^{(t+1)}\sim p_\theta(\cdot\mid x,y^{(1:t)})$ and form $y^{(1:t+1)}$
\STATE Evaluate $s\leftarrow r_\phi(x,y^{(1:t+1)})$; set $n(y^{(1:t+1)})\leftarrow 1$ and $v(y^{(1:t+1)})\leftarrow s$
\STATE Backpropagate $s$ to update $(n(\cdot),v(\cdot))$ on the path to the root
\IF{$y^{(1:t+1)}$ is terminal or $t+1=D_{\max}$}
\STATE $\mathcal{T}\leftarrow \mathcal{T}\cup\{y^{(1:t+1)}\}$; apply prefix deletion (Section~\ref{sec:wlts:diversity})
\IF{$|\mathcal{T}|\ge N$}
\STATE \textbf{break}
\ENDIF
\ELSE
\STATE Insert $y^{(1:t+1)}$ into $\mathcal{L}_{t+1}$
\ENDIF
\ENDFOR
\STATE \textbf{return} $\argmax\limits_{y\in\mathcal{T}} r_\phi(x,y)$
\end{algorithmic}
\end{algorithm}

\subsection{Diversity Enhancement}
\label{sec:wlts:diversity}
Vanilla search algorithms can over-invest in a single high-scoring branch and produce many minor variants.
WLTS encourages diversity via controllable \emph{prefix deletion}.
Fix a deletion depth $h_{\mathrm{del}}\ge 1$.
For any kept node $y^{(1:t)}$, define its truncated prefix
\[
\pi_{h_{\mathrm{del}}}\!\left(y^{(1:t)}\right)\triangleq y^{(1:\min\{h_{\mathrm{del}},t\})}.
\]
This induces \emph{families} over kept nodes: two nodes belong to the same family if they share the same $\pi_{h_{\mathrm{del}}}$.
When a terminal trajectory $y^{(1:T)}$ is found, let $p=\pi_{h_{\mathrm{del}}}(y^{(1:T)})$ and delete the remaining kept nodes in that family:
\[
\mathcal{L}_t \leftarrow \{u\in \mathcal{L}_t:\ \pi_{h_{\mathrm{del}}}(u)\neq p\},\quad \forall t.
\]
This frees width to explore alternative prefixes; setting $h_{\mathrm{del}}=1$ recovers root-level family deletion (grouping by the first generated step).

\subsection{Early Stop}
\label{sec:wlts:earlystop}
In contrast to sequence search methods, which advance multiple hypotheses in parallel, and classic tree search, which typically completes a batch of frontier nodes only after the search budget is exhausted, WLTS generates fully completed trajectories \emph{sequentially} in the order of their priority scores. This output ordering enables a natural, optional early-stopping mechanism based on agreement among terminal candidates.

For tasks whose final answers admit a canonical representation, we maintain a running frequency table over the answers extracted from completed trajectories. 
After WLTS finishes each trajectory, we canonicalize its predicted answer and update its count. If any answer reaches a predefined occurrence threshold $M$, we stop the search early and return the highest-scoring trajectory among those producing that majority answer. This strategy reduces unnecessary exploration when WLTS rapidly converges to a consistent solution, while preserving robustness when candidate answers diverge.

\subsection{Complexity Analysis}
\label{sec:wlts:complexity}
Let $G$ denote the number of generated step-level search-tree nodes for a single input instance, from initialization until WLTS returns an output.
For unconstrained tree search, the search space to depth $d$ contains $O(b^d)$ nodes under branching factor $b$, so exhaustive exploration requires $G=O(b^d)$ generated nodes.
WLTS instead uses the expandability rule to enforce $|\mathcal{L}_t|\le W$ at every depth, preventing search resources from accumulating in shallow layers.
Without prefix deletion, this width invariant gives $G \le WD_{\max}{+N}$.
With prefix deletion, let $C$ be the number of deletion operations.
Since each terminal trajectory can trigger at most one deletion operation, $C\le N$, and applying the same width bound within each deletion interval gives $G \le (C+1)WD_{\max}{+N} \le (N+1)WD_{\max}{+N}=O(NWD_{\max})$.
Thus, for fixed search-budget parameters, WLTS retains backtracking while avoiding exponential growth in generated search-tree nodes.
Appendix~\ref{app:complexity} provides the detailed accounting.

\section{Experiments}
\label{sec:experiments}

\begin{table}[t]
   \caption{Benchmark results across various datasets. Acc.(\%) and Tok.(k) report accuracy and average generator token usage (in thousands), respectively. \textbf{Average} columns report macro-averaged accuracy and tokens across the four datasets in each block; parentheses next to accuracy indicate absolute improvements over Best-of-N (in percentage points), and parentheses next to tokens indicate token usage relative to Best-of-N (= 100\%). WLTS is highlighted for readability.}
   \label{tab:combined_results}
   \centering
   \small
   \setlength{\tabcolsep}{4pt}
   \resizebox{\textwidth}{!}{%
   \begin{tabular}{lcccccccccc}
       \toprule
       \multirow{2}{*}{Classifier RM}
       & \multicolumn{2}{c}{MATH500} 
       & \multicolumn{2}{c}{OlympiadBench} 
       & \multicolumn{2}{c}{AMC23} 
       & \multicolumn{2}{c}{MBPP} 
       & \multicolumn{2}{c}{Average} \\
       \cmidrule(lr){2-3} \cmidrule(lr){4-5} \cmidrule(lr){6-7} \cmidrule(lr){8-9} \cmidrule(lr){10-11}
       & Acc.(\%) & Tok.(k) & Acc.(\%) & Tok.(k) & Acc.(\%) & Tok.(k) & Acc.(\%) & Tok.(k) & Acc.(\%) & Tok.(k) \\
       \midrule
       Llama-3.1-8B & 46.00 & -- & 16.59 & -- & 30.00 & -- & 65.90 & -- & 39.62 & -- \\
       \hspace{1em}+ Best-of-N & 66.20 & 12.68 & 28.30 & 23.16 & 50.00 & 20.24 & 72.50 & 4.42 & 54.25 & 15.13 (100\%) \\
       \hspace{1em}+ Beam Search & \underline{70.20} & 9.87 & \underline{33.33} & 14.47 & 52.50 & 13.08 & 71.70 & 4.10 & 56.93 (+2.68) & 10.38 (69\%) \\
       \hspace{1em}+ MCTS & 66.40 & 68.93 & 30.67 & 86.37 & \underline{55.00} & 91.65 & 72.50 & 13.39 & 56.14 (+1.89) & 65.09 (430\%) \\
       \hspace{1em}+ Q* & 69.80 & 57.15 & 32.74 & 60.39 & \underline{55.00} & 62.46 & \underline{73.10} & 14.22 & \underline{57.66} (+3.41) & 48.56 (321\%) \\
       \hspace{1em}+ ABMCTS-A & 66.20 & 38.16 & 30.39 & 42.48 & 52.50 & 41.34 & 70.90 & 18.39 & 55.00 (+0.75) & 35.09 (232\%) \\
       \wltsrow \hspace{1em}+ \textbf{WLTS} & \textbf{72.40} & 16.98 & \textbf{36.30} & 30.54 & \textbf{65.00} & 30.05 & \textbf{75.90} & 7.57 & \textbf{62.40} (+8.15) & 21.29 (141\%) \\
       \midrule
       \multirow{2}{*}{Generative RM} 
       & \multicolumn{2}{c}{AIME24}
       & \multicolumn{2}{c}{AIME25}
       & \multicolumn{2}{c}{ZebraLogic}
       & \multicolumn{2}{c}{LiveCodeBench}
       & \multicolumn{2}{c}{Average} \\
       \cmidrule(lr){2-3} \cmidrule(lr){4-5} \cmidrule(lr){6-7} \cmidrule(lr){8-9} \cmidrule(lr){10-11}
       & Acc.(\%) & Tok.(k) & Acc.(\%) & Tok.(k) & Acc.(\%) & Tok.(k) & Acc.(\%) & Tok.(k) & Acc.(\%) & Tok.(k) \\
       \midrule
       Qwen3-8B & 22.40 & -- & 21.07 & -- & 32.90 & -- & 23.21 & -- & 24.90 & -- \\
       \hspace{1em}+ Best-of-N & 29.33 & 60.39 & 22.67 & 50.95 & 41.10 & 36.42 & 29.52 & 9.41 & 30.66 & 39.29 (100\%) \\
       \hspace{1em}+ Beam Search & 32.67 & 72.75 & 27.33 & 55.16 & 45.80 & 52.60 & 31.93 & 16.53 & 34.43 (+3.77) & 49.26 (125\%) \\
       \hspace{1em}+ MCTS & 34.67 & 145.41 & \underline{28.67} & 121.93 & 46.50 & 107.82 & 30.12 & 67.00 & 34.99 (+4.33) & 110.54 (281\%) \\
       \hspace{1em}+ Q* & \underline{35.33} & 87.41 & 27.33 & 74.87 & \underline{46.70} & 54.70 & 31.33 & 61.75 & \underline{35.17} (+4.51) & 69.68 (177\%) \\
       \hspace{1em}+ ABMCTS-A & 32.67 & 61.69 & \textbf{29.33} & 58.88 & 40.80 & 53.66 & 29.52 & 50.76 & 33.08 (+2.42) & 56.25 (143\%) \\
       \wltsrow \hspace{1em}+ \textbf{WLTS} & \textbf{42.67} & 42.91 & \textbf{29.33} & 38.21 & \textbf{48.40} & 36.41 & \textbf{33.13} & 14.16 & \textbf{38.38} (+7.72) & 32.92 (84\%) \\
       \midrule
      Qwen3-14B & 28.27 & -- & 25.07 & -- & 44.90 & -- & 27.71 & -- & 31.49 & -- \\
      \hspace{1em}+ Best-of-N & 40.67 & 56.16 & 32.00 & 46.06 & 50.60 & 34.22 & 31.93 & 9.09 & 38.80 & 36.38 (100\%) \\
      \hspace{1em}+ Beam Search & 40.00 & 76.74 & 33.33 & 62.52 & \underline{55.10} & 33.84 & 34.34 & 18.55 & 40.70 (+1.90) & 47.91 (132\%) \\
      \hspace{1em}+ MCTS & \underline{42.00} & 155.00 & 33.33 & 152.75 & 54.40 & 112.89 & 34.94 & 40.98 & \underline{41.17} (+2.37) & 115.41 (317\%) \\
      \hspace{1em}+ Q* & 38.00 & 57.28 & \underline{33.66} & 51.57 & 51.40 & 62.38 & \textbf{38.55} & 43.42 & 40.40 (+1.60) & 53.66 (147\%) \\
      \hspace{1em}+ ABMCTS-A & 40.67 & 60.20 & 31.33 & 59.93 & 52.20 & 53.61 & 37.95 & 42.53 & 40.54 (+1.74) & 54.07 (149\%) \\
      \wltsrow \hspace{1em}+ \textbf{WLTS} & \textbf{46.00} & 32.58 & \textbf{35.33} & 33.70 & \textbf{63.60} & 37.69 & \underline{37.99} & 14.64 & \textbf{45.73} (+6.93) & 29.65 (82\%) \\
       \bottomrule
   \end{tabular}%
   }
   \vspace{-1em}
\end{table}

\subsection{Experiment Setup}

To more comprehensively evaluate WLTS, we conduct experiments using multiple representative model configurations.
For the classifier reward model, we use Llama-3.1-8B-Instruct as the base model and Skywork-o1-Open-PRM-Qwen-2.5-7B as the reward model~\citep{SkyworkReward2024}.
For the generative reward model, we use Qwen3-8B and Qwen3-14B as the base models, and use them as their own verifiers, as they possess the ability to make effective judgments.
Given the strength of the base models, we select challenging reasoning tasks that cover both mathematical and coding domains to comprehensively evaluate WLTS.
For Llama-3.1-8B-Instruct, we use math problems from MATH500, Olympiad Bench, AMC23, and coding problems from MBPP.
For Qwen3-8B and Qwen3-14B, we use math problems from AIME24 and AIME25, coding problems from LiveCodeBench, and the logical-reasoning benchmark ZebraLogic.
For sequence search, we use best-of-N and beam search as representative baselines.
For tree search, we include MCTS, as well as Q*~\citep{Qstar2024} and ABMCTS-A~\citep{WiderDeeper2025}, two recent efficiency-oriented tree-search variants.

\subsection{Main Results}

\textbf{Evaluation protocol.} Following prior test-time scaling evaluations~\citep{OptimalScaling2024,SelfVerify2025,lifshitz2025mav}, we use a \emph{candidate-count-matched protocol}, where each method targets up to $N{=}16$ complete candidate answers per problem.
Best-of-N and Beam Search directly decode $N$ complete trajectories, whereas MCTS, Q*, and ABMCTS-A search over step-level prefixes under fixed expansion budgets and, when fewer than $N$ full trajectories are produced, complete selected prefixes with a single forward pass following the common budget-limited tree-search completion protocol~\citep{Qstar2024}.
WLTS uses the same candidate target and may terminate earlier if the early stopping condition is met.
Because matched candidate counts do not imply matched token costs, we report total generated tokens for the full search procedure, including intermediate branches, discarded candidates, and final completions; Appendix~\ref{sec:appendix:token_matched} provides a token-budget-matched comparison.

\textbf{Results.} Table~\ref{tab:combined_results} shows that inference-time search improves over single-pass decoding across all model and verifier settings, and step-level search generally outperforms trajectory-level Best-of-N.
However, existing tree-search baselines offer no compelling advantage over Beam Search once their substantially higher token usage is considered.
The key reason is that, in long-CoT reasoning, finite search budgets are often spent disproportionately on shallow regions, leaving deeper reasoning steps without sustained verifier supervision and forcing completion from the current best prefix, which degrades final answer quality.
Although Q* and ABMCTS-A reduce token usage by limiting the number of children expanded at each step, under fixed expansion budgets they do not yield a substantial accuracy advantage over standard MCTS.
In contrast, WLTS is designed for this long-CoT setting: it inherits the linearly constrained search space of sequence search, distributing budget more evenly across the full reasoning process, while preserving the recovery structure of tree search so that backpropagation can return to more promising prefixes after errors.
This recovery behavior is empirically important: with Qwen3-14B on AIME24 and AIME25, 59 of 60 solved AIME24 instances and all 51 solved AIME25 instances are solved only after at least one backtracking event; moreover, 10 and 9 solved instances, respectively, are solved only after WLTS first reaches an incorrect terminal answer and then recovers through backtracking, providing direct evidence of recovery behavior unavailable to beam-style search.

\subsection{Components Ablation}
\label{sec:exp:components}

\begin{table}[t]
   \caption{Component-wise ablation of WLTS on Qwen3-14B. Each cell reports accuracy (\%) / generated tokens (k) per problem. PD = prefix deletion; ES = agreement-based early stopping. The Beam Search and ABMCTS-A rows reproduce Table~\ref{tab:combined_results} for reference.}
   \label{tab:components_ablation}
   \centering
   \small
   \setlength{\tabcolsep}{4pt}
   \begin{tabular}{lccccc}
      \toprule
      Method & AIME24 & AIME25 & LiveCodeBench & ZebraLogic & Average \\
      \midrule
      Beam Search & 40.00 / 76.74  & 33.33 / 62.52  & 34.34 / 18.55 & 55.10 / 33.84  & 40.70 / 47.91  \\
      ABMCTS-A    & 40.67 / 60.20  & 31.33 / 59.93  & \underline{37.95} / 42.53 & 52.20 / 53.61  & 40.54 / 54.07  \\
      \midrule
      \wltsrow Core WLTS                    & 41.33 / 27.28 & 32.33 / 25.88 & 37.35 / 4.09 & 55.80 / 26.51 & 41.70 / 20.94 \\
      \wltsrow Core WLTS + ES               & 42.67 / 19.72 & 31.33 / 18.84 & 37.35 / 4.09 & \underline{57.50} / 11.85         & 42.21 / 13.63 \\
      \wltsrow Core WLTS + PD               & \textbf{47.33} / 50.01 & \textbf{36.67} / 52.45 & \textbf{37.99} / 14.64 & \underline{57.50} / 54.97 & \underline{44.87} / 43.02 \\
      \wltsrow \textbf{WLTS} & \underline{46.00} / 32.58 & \underline{35.33} / 33.70 & \textbf{37.99} / 14.64 & \textbf{63.60} / 37.69 & \textbf{45.73} / 29.65 \\
      \bottomrule
   \end{tabular}%
   \vspace{-0.8em}
\end{table}

WLTS combines three elements: (i) the core width-limited backtracking structure, (ii) prefix deletion (PD) for diversity, and (iii) agreement-based early stopping (ES). To disentangle their contributions, we ablate the optional components on Qwen3-14B and report accuracy and generated-token usage across AIME24, AIME25, LiveCodeBench, and ZebraLogic. We also include Beam Search and ABMCTS-A from Table~\ref{tab:combined_results} as reference baselines, since they provide the strongest accuracy--token trade-offs among non-WLTS methods in this setting.

Table~\ref{tab:components_ablation} shows three main patterns. First, \emph{Core WLTS}, with neither PD nor ES, already outperforms Beam Search and ABMCTS-A on average while using substantially fewer tokens, indicating that the main accuracy--efficiency gain comes from the core width-limited backtracking structure rather than optional deployment components. Its per-layer cap distributes search resources across the full response length, allowing each step toward the final answer to receive effective verifier supervision, while backtracking stops unpromising intermediate trajectories and redirects capacity to alternative candidate nodes. Second, ES substantially reduces token usage on canonicalizable-answer benchmarks and can also improve accuracy, since recurring answers across completed trajectories are more likely to be correct. This matches its role as an efficiency mechanism for canonicalizable outputs. Third, PD injects exploration diversity by deleting the prefix family of a completed trajectory, freeing width for alternative high-level prefixes and substantially improving accuracy at higher token cost. This supports our expectation that prefix deletion raises the attainable performance ceiling by widening the explored prefix space. Combining ES and PD achieves the best average accuracy while reducing tokens relative to Core WLTS + PD, showing that they are complementary: PD broadens the explored prefix space, whereas ES truncates search once the widened exploration converges on a consistent answer. Overall, ES and PD serve as deployment-time knobs for different accuracy--efficiency requirements. We further analyze prefix-deletion depth $h_{\mathrm{del}}$ and agreement threshold $M$ in Appendices~\ref{sec:appendix:diversity_details} and~\ref{sec:appendix:early_stopping}.

\subsection{Inference-time Scaling Law}
\label{sec:exp:scaling}

\begin{wrapfigure}{r}{0.5\textwidth}
   \vspace{-1.em}
   \centering
   \includegraphics[width=0.48\textwidth]{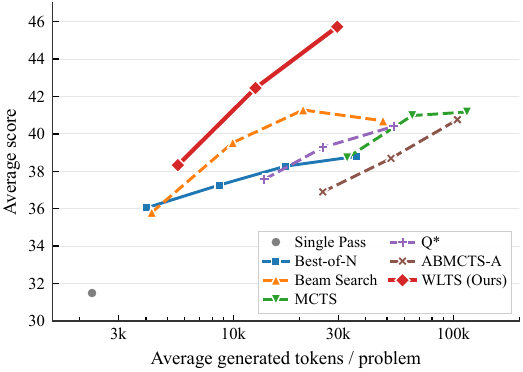}
   \caption{Inference-time scaling on Qwen3-14B.}
   \label{fig:inference_scaling}
   \vspace{-1.em}
\end{wrapfigure}

We assess inference-time compute scaling by varying each method's search budget and recording macro-averaged accuracy and generated-token usage.
The experiment uses Qwen3-14B, with results macro-averaged over AIME24, AIME25, ZebraLogic, and LiveCodeBench; per-benchmark results are provided in Appendix~\ref{sec:appendix:qwen14b_details} (Table~\ref{tab:qwen14b_details}).
As shown in Figure~\ref{fig:inference_scaling}, all search algorithms generally improve with additional tokens, but WLTS scales most strongly.
Increasing the WLTS budget expands exploration under a strict per-layer width cap, so additional computation is distributed across reasoning depths and yields more high-quality completed candidates.
In contrast, for tree-search methods such as MCTS, Q*, and ABMCTS-A, additional budget can still be diluted by shallow nodes, leaving less effective search effort for deeper states and producing flatter scaling curves.
Beam Search initially improves sharply with larger candidate budgets but eventually bottlenecks: in long-CoT reasoning, retained candidates can concentrate around superficially plausible prefix families, and irreversible pruning prevents recovery from such local optima.
This failure mode motivates preserving backtracking in WLTS.\looseness=-1

\subsection{Verifier Sensitivity}
\label{sec:exp:verifier}

\begin{table}[t]
   \caption{Verifier sensitivity on Llama-3.1-8B. The four benchmark columns and ``Gen.\ Avg'' report accuracy (\%) under a same-backbone generative verifier implemented by Llama-3.1-8B itself. ``Cls.\ Avg'' shows the corresponding Skywork-PRM macro-average from Table~\ref{tab:combined_results} for direct contrast.}
   \label{tab:verifier_sensitivity}
   \centering
   \small
   \setlength{\tabcolsep}{6pt}
   \begin{tabular}{lcccccc}
      \toprule
      Method & MATH500 & Olympiad Bench & AMC23 & MBPP & Gen.\ Avg & Cls.\ Avg \\
      \midrule
      Best-of-N   & 49.80 & 19.41 & 25.75 & 66.00 & 40.24 & 54.25 \\
      Beam Search & 50.00 & 18.96 & \textbf{33.25} & \underline{67.60} & \underline{42.45} & 56.93 \\
      MCTS        & 51.40 & 18.82 & 31.25 & 66.80 & 42.07 & 56.14 \\
      Q*          & \underline{51.80} & \underline{20.15} & 23.75 & 66.40 & 40.53 & \underline{57.66} \\
      ABMCTS-A    & 50.40 & 18.37 & 30.00 & 66.70 & 41.37 & 55.00 \\
      \midrule
      \wltsrow \textbf{WLTS} & \textbf{57.60} & \textbf{22.07} & \underline{32.50} & \textbf{68.30} & \textbf{45.12} & \textbf{62.40} \\
      \bottomrule
   \end{tabular}
   \vspace{-0.8em}
\end{table}

In the main experiments, each model setting is paired with a strong corresponding verifier: Llama-3.1-8B uses the specialized Skywork-PRM classifier reward model, while the Qwen3 models use their stronger thinking-mode generative verification.
Here, we stress-test robustness under a weaker verifier by using Llama-3.1-8B itself as a generative verifier, whose judgments are substantially weaker than Skywork-PRM.
Table~\ref{tab:verifier_sensitivity} shows that, as expected, degrading the verifier from the specialized PRM to the self-verifier substantially lowers absolute scores for all methods.
Nevertheless, WLTS retains the best macro-average accuracy, ranking first on three of the four tasks and second on the remaining one.
This suggests that WLTS is not tied to a particular verifier strength: even with weaker step-level judgments, its width-limited backtracking structure remains robust.

\section{Related Work}

\subsection{Inference-Time Scaling via Sequence Search}
Beyond policy optimization with verifiable rewards~\citep{DeepSeekR12025,NOPO2026}, LLM reasoning can also be improved by scaling inference-time search, which can rival parameter scaling under fixed compute budgets~\citep{OptimalScaling2024,SearchSurvey2025,TreeSearchSurvey2025}.
The simplest \emph{sequence search} methods sample multiple complete solutions and select an output by reward-model scoring or agreement.
Best-of-$N$ returns the reward-model top-ranked solution, with recent work studying optimal test-time compute allocation~\citep{OptimalScaling2024}.
Self-consistency selects the majority answer across chain-of-thought samples, improving robustness to stochastic generation~\citep{wang2022selfconsistency}.
These methods are simple and scale linearly in the number of sampled trajectories, but operate only over full solutions and cannot revise erroneous prefixes once committed.

More structured sequence search scores partial prefixes.
Beam search maintains a fixed-size frontier at each step and can incorporate task-specific or learned scoring signals~\citep{sutskever2014seq2seq}.
REBASE can be viewed as a reward-guided extension of beam search, replacing uniform expansion counts per prefix with allocations based on verifier scores~\citep{REBASE2025}.
However, pruning in these methods is irreversible: if the frontier collapses onto an incorrect reasoning line early, later evidence cannot recover discarded alternative prefixes.
Diversity-promoting decoding methods reduce hypothesis correlation but still remain purely forward decoding methods~\citep{li2016diverse}.
WLTS is complementary: it preserves tree-search-style backtracking while using prefix deletion to remove over-explored prefix families after each terminal trajectory, encouraging exploration of distinct high-level plans.

\subsection{Tree Search for LLM Reasoning}
To enable revision and backtracking, recent methods cast multi-step reasoning as tree search over partial trajectories, such as ToT~\citep{yao2023tot} and GoT~\citep{besta2024graphthoughts}, often with step-level feedback from process reward models for intermediate correctness~\citep{lightman2023verify}.
As strong LLMs increasingly serve as generative verifiers, signal quality and granularity have become key determinants of search effectiveness~\citep{Heimdall2025,GenerativeVerifiers2025}.
Compute allocation is central: value-guided pruning/prioritization focus expansions on promising prefixes, while uncertainty-based criteria improve exploration under noisy scores~\citep{UncertaintyGuidedOptimization2024,UncertaintyOfThoughts2024}.
Planning-oriented methods such as Q* and rStar-Math highlight the gains from exploring and revising intermediate decisions under fixed budgets~\citep{Qstar2024,rStarMath2025}.
Recent work improves the practicality and budget efficiency of LLM tree search~\citep{DonGetLost2025,liteSearch2024,ETS2025}, including adaptive branching width--depth control~\citep{WiderDeeper2025} and decisions about when to branch~\citep{ChainInTree2026}.

However, adaptive branching control does not by itself enforce a strict per-layer capacity limit and can still allow rapid frontier growth with depth.
WLTS shares the goal of budgeted backtracking but differs structurally by enforcing a strict \emph{per-layer} width cap via a dynamic expandability criterion, guaranteeing linear-in-depth growth of generated search-tree nodes.
This constraint is orthogonal to the reward model and can be combined with value- or uncertainty-guided selection, pruning, and diversification techniques.

\section{Conclusion and Limitations}

As LLM responses grow longer, existing tree-search algorithms struggle to distribute limited compute across reasoning depths, with much of the budget consumed by shallow exploration.
Motivated by this challenge in long-chain-of-thought reasoning, we presented \emph{Width-Limited Tree Search} (WLTS), an LLM inference search algorithm that reconciles the strengths of sequence search and tree search.
WLTS preserves backtracking while provably keeping generated search-tree growth linear in reasoning depth through a strict per-layer width cap enforced by a dynamic expandability criterion.
This width-limited structure distributes search budget across the full response length and sustains verifier supervision throughout the final-answer generation process.
To accommodate different accuracy--token trade-offs, WLTS further supports two optional components: prefix deletion for exploration diversity and agreement-based early stopping for reducing computation on canonicalizable-output tasks.
Extensive experiments consistently demonstrate the effectiveness, generality, and robustness of our method.\looseness=-1

\paragraph{Limitations.}
The current strict width cap remains a relatively static capacity-control mechanism.
As LLM responses grow longer, low-quality prefixes may occupy limited per-layer capacity before sufficient evidence emerges to identify them and reallocate capacity, thereby reducing the effective backtracking space for recovery from early mistakes.
A promising direction is to develop more proactive control mechanisms, such as dynamically releasing unpromising nodes during exploration or adaptively adjusting the per-layer width.

{
\small
\bibliographystyle{plainnat}
\begingroup
\let\plainbibitem\bibitem
\renewcommand{\bibitem}[2][]{\color{black}\plainbibitem[#1]{#2}}
\bibliography{references}

\begin{thebibliography}{35}
\providecommand{\natexlab}[1]{#1}
\providecommand{\url}[1]{\texttt{#1}}
\expandafter\ifx\csname urlstyle\endcsname\relax
  \providecommand{\doi}[1]{doi: #1}\else
  \providecommand{\doi}{doi: \begingroup \urlstyle{rm}\Url}\fi

\bibitem[Besta et~al.(2024)Besta, Blach, Kubicek, Gerstenberger, Podstawski, Gianinazzi, Gajda, Lehmann, Niewiadomski, Nyczyk, and Hoefler]{besta2024graphthoughts}
Maciej Besta, Nils Blach, Ales Kubicek, Robert Gerstenberger, Michal Podstawski, Lukas Gianinazzi, Joanna Gajda, Tomasz Lehmann, Hubert Niewiadomski, Piotr Nyczyk, and Torsten Hoefler.
\newblock Graph of thoughts: Solving elaborate problems with large language models.
\newblock In \emph{AAAI Conference on Artificial Intelligence (AAAI)}, pages 17682--17690, 2024.

\bibitem[Browne et~al.(2012)Browne, Powley, Whitehouse, Lucas, Cowling, Rohlfshagen, Tavener, Perez, Samothrakis, and Colton]{browne2012mcts}
Cameron Browne, Edward Powley, Daniel Whitehouse, Simon~M. Lucas, Peter~I. Cowling, Philipp Rohlfshagen, Stephen Tavener, Diego Perez, Spyridon Samothrakis, and Simon Colton.
\newblock A survey of {Monte Carlo} tree search methods.
\newblock \emph{IEEE Transactions on Computational Intelligence and AI in Games}, 4\penalty0 (1):\penalty0 1--43, 2012.

\bibitem[Grosse et~al.(2024)Grosse, Wu, Rashid, Zhang, Hennig, Poupart, and Kristiadi]{UncertaintyGuidedOptimization2024}
Julia Grosse, Ruotian Wu, Ahmad Rashid, Cheng Zhang, Philipp Hennig, Pascal Poupart, and Agustinus Kristiadi.
\newblock Uncertainty-guided likelihood tree search.
\newblock \emph{arXiv preprint arXiv:2407.03951}, 2024.

\bibitem[Guan et~al.(2025)Guan, Zhang, Liu, Shang, Sun, Zhu, Yang, and Yang]{rStarMath2025}
Xinyu Guan, Li~Lyna Zhang, Yifei Liu, Ning Shang, Youran Sun, Yi~Zhu, Fan Yang, and Mao Yang.
\newblock {rStar-Math}: Small {LLMs} can master math reasoning with self-evolved deep thinking.
\newblock In \emph{International Conference on Machine Learning (ICML)}, 2025.

\bibitem[Guo et~al.(2025)Guo, Yang, Zhang, et~al.]{DeepSeekR12025}
Daya Guo, Dejian Yang, Haowei Zhang, et~al.
\newblock {DeepSeek-R1} incentivizes reasoning in {LLMs} through reinforcement learning.
\newblock \emph{Nature}, 645:\penalty0 633--638, 2025.

\bibitem[Hooper et~al.(2025)Hooper, Kim, Moon, Dilmen, Maheswaran, Lee, Mahoney, Shao, Keutzer, and Gholami]{ETS2025}
Coleman Hooper, Sehoon Kim, Suhong Moon, Kerem Dilmen, Monishwaran Maheswaran, Nicholas Lee, Michael~W. Mahoney, Yakun~Sophia Shao, Kurt Keutzer, and Amir Gholami.
\newblock {ETS}: Efficient tree search for inference-time scaling.
\newblock \emph{arXiv preprint arXiv:2502.13575}, 2025.

\bibitem[Hu et~al.(2024)Hu, Liu, Feng, Zhao, Ng, Luu, He, Koh, and Hooi]{UncertaintyOfThoughts2024}
Zhiyuan Hu, Chumin Liu, Xidong Feng, Yilun Zhao, See{-}Kiong Ng, Anh~Tuan Luu, Junxian He, Pang~Wei Koh, and Bryan Hooi.
\newblock Uncertainty of thoughts: Uncertainty-aware planning enhances information seeking in large language models.
\newblock In \emph{Advances in Neural Information Processing Systems (NeurIPS)}, 2024.

\bibitem[Inoue et~al.(2025)Inoue, Misaki, Imajuku, Kuroki, Nakamura, and Akiba]{WiderDeeper2025}
Yuichi Inoue, Kou Misaki, Yuki Imajuku, So~Kuroki, Taishi Nakamura, and Takuya Akiba.
\newblock Wider or deeper? {S}caling {LLM} inference-time compute with adaptive branching tree search.
\newblock In \emph{Advances in Neural Information Processing Systems (NeurIPS)}, 2025.

\bibitem[Jiang et~al.(2024)Jiang, Chen, Min, Chen, Cheng, Wang, Tang, Sun, Deng, Zhao, Liu, Yan, Xie, Wang, and Wen]{RewardGuidedTreeSearch2024}
Jinhao Jiang, Zhipeng Chen, Yingqian Min, Jie Chen, Xiaoxue Cheng, Jiapeng Wang, Yiru Tang, Haoxiang Sun, Jia Deng, Wayne~Xin Zhao, Zheng Liu, Dong Yan, Jian Xie, Zhongyuan Wang, and Ji{-}Rong Wen.
\newblock Enhancing {LLM} reasoning with reward-guided tree search.
\newblock \emph{arXiv preprint arXiv:2411.11694}, 2024.

\bibitem[Kocsis and Szepesv{\'{a}}ri(2006)]{kocsis2006uct}
Levente Kocsis and Csaba Szepesv{\'{a}}ri.
\newblock Bandit based {Monte-Carlo} planning.
\newblock In \emph{European Conference on Machine Learning (ECML)}, pages 282--293, Berlin, 2006.

\bibitem[Li and Jurafsky(2016)]{li2016diverse}
Jiwei Li and Dan Jurafsky.
\newblock Mutual information and diverse decoding improve neural machine translation.
\newblock \emph{arXiv preprint arXiv:1601.00372}, 2016.

\bibitem[Li(2025)]{SearchSurvey2025}
Xinzhe Li.
\newblock A survey on {LLM} test-time compute via search: Tasks, {LLM} profiling, search algorithms, and relevant frameworks.
\newblock \emph{Transactions on Machine Learning Research}, 2025.

\bibitem[Li(2026)]{ChainInTree2026}
Xinzhe Li.
\newblock {Chain-in-Tree}: Back to sequential reasoning in {LLM} tree search.
\newblock In \emph{Findings of the Association for Computational Linguistics (ACL)}, pages 4365--4392, 2026.

\bibitem[Lifshitz et~al.(2025)Lifshitz, McIlraith, and Du]{lifshitz2025mav}
Shalev Lifshitz, Sheila~A. McIlraith, and Yilun Du.
\newblock Multi-agent verification: Scaling test-time compute with multiple verifiers.
\newblock In \emph{Conference on Language Modeling (COLM)}, 2025.

\bibitem[Lightman et~al.(2024)Lightman, Kosaraju, Burda, Edwards, Baker, Lee, Leike, Schulman, Sutskever, and Cobbe]{lightman2023verify}
Hunter Lightman, Vineet Kosaraju, Yuri Burda, Harrison Edwards, Bowen Baker, Teddy Lee, Jan Leike, John Schulman, Ilya Sutskever, and Karl Cobbe.
\newblock Let's verify step by step.
\newblock In \emph{International Conference on Learning Representations (ICLR)}, 2024.

\bibitem[Liu et~al.(2024)Liu, Zeng, Liu, Yan, He, Wang, Yan, Liu, and Zhou]{SkyworkReward2024}
Chris~Yuhao Liu, Liang Zeng, Jiacai Liu, Rui Yan, Jujie He, Chaojie Wang, Shuicheng Yan, Yang Liu, and Yahui Zhou.
\newblock {Skywork-Reward}: Bag of tricks for reward modeling in {LLMs}.
\newblock \emph{arXiv preprint arXiv:2410.18451}, 2024.

\bibitem[{OpenAI}(2023)]{openai2023gpt4}
{OpenAI}.
\newblock {GPT-4} technical report.
\newblock \emph{arXiv preprint arXiv:2303.08774}, 2023.

\bibitem[Pang et~al.(2024)Pang, Wang, Li, Chen, Xu, Zhang, and Yu]{RLContemplation2024}
Jing{-}Cheng Pang, Pengyuan Wang, Kaiyuan Li, Xiong{-}Hui Chen, Jiacheng Xu, Zongzhang Zhang, and Yang Yu.
\newblock Language model self-improvement by reinforcement learning contemplation.
\newblock In \emph{International Conference on Learning Representations (ICLR)}, 2024.

\bibitem[Shi and Jin(2025)]{Heimdall2025}
Wenlei Shi and Xing Jin.
\newblock Heimdall: Test-time scaling on the generative verification.
\newblock \emph{arXiv preprint arXiv:2504.10337}, 2025.

\bibitem[Snell et~al.(2025)Snell, Lee, Xu, and Kumar]{OptimalScaling2024}
Charlie Snell, Jaehoon Lee, Kelvin Xu, and Aviral Kumar.
\newblock Scaling {LLM} test-time compute optimally can be more effective than scaling model parameters.
\newblock In \emph{International Conference on Learning Representations (ICLR)}, 2025.

\bibitem[Sutskever et~al.(2014)Sutskever, Vinyals, and Le]{sutskever2014seq2seq}
Ilya Sutskever, Oriol Vinyals, and Quoc~V. Le.
\newblock Sequence to sequence learning with neural networks.
\newblock In \emph{Advances in Neural Information Processing Systems (NeurIPS)}, pages 3104--3112, 2014.

\bibitem[Wang et~al.(2025{\natexlab{a}})Wang, Song, Tian, Peng, Yu, Mi, Su, and Yu]{liteSearch2024}
Ante Wang, Linfeng Song, Ye~Tian, Baolin Peng, Dian Yu, Haitao Mi, Jinsong Su, and Dong Yu.
\newblock {LiteSearch}: Efficient tree search with dynamic exploration budget for math reasoning.
\newblock In \emph{AAAI Conference on Artificial Intelligence (AAAI)}, pages 25318--25326, 2025{\natexlab{a}}.

\bibitem[Wang et~al.(2025{\natexlab{b}})Wang, Song, Tian, Yu, Mi, Duan, Tu, Su, and Yu]{DonGetLost2025}
Ante Wang, Linfeng Song, Ye~Tian, Dian Yu, Haitao Mi, Xiangyu Duan, Zhaopeng Tu, Jinsong Su, and Dong Yu.
\newblock Don't get lost in the trees: Streamlining {LLM} reasoning by overcoming tree search exploration pitfalls.
\newblock In \emph{Annual Meeting of the Association for Computational Linguistics (ACL)}, pages 23946--23959, 2025{\natexlab{b}}.

\bibitem[Wang et~al.(2024{\natexlab{a}})Wang, Deng, Lyu, Zeng, He, Yan, and An]{Qstar2024}
Chaojie Wang, Yanchen Deng, Zhiyi Lyu, Liang Zeng, Jujie He, Shuicheng Yan, and Bo~An.
\newblock {Q*}: Improving multi-step reasoning for {LLMs} with deliberative planning.
\newblock \emph{arXiv preprint arXiv:2406.14283}, 2024{\natexlab{a}}.

\bibitem[Wang et~al.(2024{\natexlab{b}})Wang, Li, Shao, Xu, Dai, Li, Chen, Wu, and Sui]{MathShepherd2024}
Peiyi Wang, Lei Li, Zhihong Shao, Runxin Xu, Damai Dai, Yifei Li, Deli Chen, Yu~Wu, and Zhifang Sui.
\newblock {Math-Shepherd}: Verify and reinforce {LLMs} step-by-step without human annotations.
\newblock In \emph{Annual Meeting of the Association for Computational Linguistics (ACL)}, pages 9426--9439, 2024{\natexlab{b}}.

\bibitem[Wang et~al.(2023)Wang, Wei, Schuurmans, Le, Chi, Narang, Chowdhery, and Zhou]{wang2022selfconsistency}
Xuezhi Wang, Jason Wei, Dale Schuurmans, Quoc~V. Le, Ed~H. Chi, Sharan Narang, Aakanksha Chowdhery, and Denny Zhou.
\newblock Self-consistency improves chain of thought reasoning in language models.
\newblock In \emph{International Conference on Learning Representations (ICLR)}, 2023.

\bibitem[Wei et~al.(2022)Wei, Wang, Schuurmans, Bosma, Ichter, Xia, Chi, Le, and Zhou]{wei2022cot}
Jason Wei, Xuezhi Wang, Dale Schuurmans, Maarten Bosma, Brian Ichter, Fei Xia, Ed~H. Chi, Quoc~V. Le, and Denny Zhou.
\newblock Chain-of-thought prompting elicits reasoning in large language models.
\newblock In \emph{Advances in Neural Information Processing Systems (NeurIPS)}, 2022.

\bibitem[Wei et~al.(2025)Wei, Zhang, Yang, Huang, Cao, Xu, Zhuang, Gao, Abdul{-}Mageed, Lakshmanan, You, Ouyang, and Sun]{TreeSearchSurvey2025}
Jiaqi Wei, Xiang Zhang, Yuejin Yang, Wenxuan Huang, Juntai Cao, Sheng Xu, Xiang Zhuang, Zhangyang Gao, Muhammad Abdul{-}Mageed, Laks V.~S. Lakshmanan, Chenyu You, Wanli Ouyang, and Siqi Sun.
\newblock Unifying tree search algorithm and reward design for {LLM} reasoning: {A} survey.
\newblock \emph{arXiv preprint arXiv:2510.09988}, 2025.

\bibitem[Wu et~al.(2025)Wu, Sun, Li, Welleck, and Yang]{REBASE2025}
Yangzhen Wu, Zhiqing Sun, Shanda Li, Sean Welleck, and Yiming Yang.
\newblock Inference scaling laws: An empirical analysis of compute-optimal inference for {LLM} problem-solving.
\newblock In \emph{International Conference on Learning Representations (ICLR)}, 2025.

\bibitem[Xu et~al.(2026{\natexlab{a}})Xu, Chen, Xu, and An]{ERPO2026}
Jiacheng Xu, Feng Chen, Xiuneng Xu, and Bo~An.
\newblock Entropy-regularized rank-masked policy optimization for test-time reinforcement learning in code generation.
\newblock In \emph{Conference on Empirical Methods in Natural Language Processing (EMNLP)}, pages 47843--47857, Budapest, Hungary, 2026{\natexlab{a}}.

\bibitem[Xu et~al.(2026{\natexlab{b}})Xu, He, Zhang, Wang, and An]{NOPO2026}
Jiacheng Xu, Shuo He, Fuxiang Zhang, Chaojie Wang, and Bo~An.
\newblock Negative-only policy optimization for one-sided verifiable rewards.
\newblock \emph{arXiv preprint arXiv:2611.17287}, 2026{\natexlab{b}}.

\bibitem[Xu et~al.(2026{\natexlab{c}})Xu, Zhang, Lyu, Zhang, Wang, Liu, and An]{TCS2026}
Jiacheng Xu, Wentao Zhang, Zhiyi Lyu, Fuxiang Zhang, Chaojie Wang, Yang Liu, and Bo~An.
\newblock Two-stage reinforcement learning for sound and adversarial test generation in code {LLM}s.
\newblock In \emph{Findings of the Association for Computational Linguistics (EMNLP)}, pages 37031--37051, Budapest, Hungary, 2026{\natexlab{c}}.

\bibitem[Yao et~al.(2023)Yao, Yu, Zhao, Shafran, Griffiths, Cao, and Narasimhan]{yao2023tot}
Shunyu Yao, Dian Yu, Jeffrey Zhao, Izhak Shafran, Thomas~L. Griffiths, Yuan Cao, and Karthik Narasimhan.
\newblock Tree of thoughts: Deliberate problem solving with large language models.
\newblock In \emph{Advances in Neural Information Processing Systems (NeurIPS)}, 2023.

\bibitem[Zhang et~al.(2025{\natexlab{a}})Zhang, Xu, Wang, Cui, Liu, and An]{SelfVerify2025}
Fuxiang Zhang, Jiacheng Xu, Chaojie Wang, Ce~Cui, Yang Liu, and Bo~An.
\newblock Incentivizing {LLMs} to self-verify their answers.
\newblock In \emph{Advances in Neural Information Processing Systems (NeurIPS)}, 2025{\natexlab{a}}.

\bibitem[Zhang et~al.(2025{\natexlab{b}})Zhang, Hosseini, Bansal, Kazemi, Kumar, and Agarwal]{GenerativeVerifiers2025}
Lunjun Zhang, Arian Hosseini, Hritik Bansal, Mehran Kazemi, Aviral Kumar, and Rishabh Agarwal.
\newblock Generative verifiers: Reward modeling as next-token prediction.
\newblock In \emph{International Conference on Learning Representations (ICLR)}, 2025{\natexlab{b}}.

\end{thebibliography}
\endgroup
}

\appendix

\newpage
\section{Detailed Complexity Accounting}
\label{app:complexity}

This appendix gives a more explicit accounting of the generated search tree discussed in Section~\ref{sec:wlts:complexity}.
We count search-tree nodes generated for a single input instance, from initialization until WLTS returns an output, including nodes later removed by prefix deletion.
This accounting concerns the size of the generated search tree; empirical accuracy--cost tradeoffs are evaluated in Section~\ref{sec:experiments}.

For unconstrained tree search, the search space to depth $d$ contains $O(b^d)$ nodes under branching factor $b$, so exhaustive exploration requires $O(b^d)$ generated nodes.
WLTS avoids this exponential depth dependence through the per-layer width cap.

\begin{lemma}[Width invariant]
\label{lem:wlts-width}
Throughout WLTS, each layer contains at most $W$ kept search nodes:
\[
|\mathcal{L}_t|\le W,\quad \forall t.
\]
\end{lemma}

\begin{proof}
The claim holds at initialization.
Consider one expansion that selects a node at depth $t\ge t_{\mathrm{full}}$.
The generated child is associated with depth $t+1$; if it remains expandable, it is inserted into layer $\mathcal{L}_{t+1}$.
By definition of $t_{\mathrm{full}}$, every layer strictly deeper than $t_{\mathrm{full}}$ is not saturated.
Since $t+1\ge t_{\mathrm{full}}+1$, any such insertion preserves $|\mathcal{L}_{t+1}|\le W$.
All other layers are unchanged, and prefix deletion only removes kept nodes.
Therefore, the width cap is preserved after each expansion.
\end{proof}

Each loop iteration of WLTS generates exactly one step-level node, which is either inserted into a layer $\mathcal{L}_{t+1}$ (\emph{kept}) or added to the terminal set $\mathcal{T}$ (\emph{terminal}).
The two corollaries below bound these contributions separately.

\begin{corollary}[Linear-in-depth search-space growth of Core WLTS]
\label{cor:wlts-core-linear}
Consider Core WLTS without prefix deletion.
Let $D_{\max}$ be the maximum reasoning depth, $W$ the per-layer width cap, and $N$ the terminal-candidate target.
Then the total number of search-tree nodes generated for a single input instance is bounded by
\[
G_{\mathrm{core}} \le WD_{\max}+N.
\]
Thus, for fixed $W$ and $N$, Core WLTS has linear search-space growth in reasoning depth.
\end{corollary}

\begin{proof}
Each generated node is either kept or terminal.
Without deletion, no layer capacity is reopened after a kept node is generated at that depth.
By Lemma~\ref{lem:wlts-width}, each layer can therefore receive at most $W$ kept generated nodes, so across depths $1,\ldots,D_{\max}$ the total number of kept generated nodes is at most $WD_{\max}$.
The number of terminal nodes is bounded by $|\mathcal{T}|\le N$ at termination, since the loop breaks once $|\mathcal{T}|\ge N$.
Summing the two contributions gives $G_{\mathrm{core}}\le WD_{\max}+N$.
\end{proof}

\begin{corollary}[Linear-in-depth search-space growth with deletion]
\label{cor:wlts-deletion-linear}
Consider WLTS with prefix deletion.
Let $C$ be the number of deletion operations during this invocation.
Then the total number of generated search-tree nodes is bounded by
\[
G_{\mathrm{del}} \le (C+1)WD_{\max}+N.
\]
Since prefix deletion is triggered only after a terminal trajectory is found and WLTS collects at most $N$ terminal trajectories, $C\le N$.
Therefore,
\[
G_{\mathrm{del}}
\le
(N+1)WD_{\max}+N
=
O(NWD_{\max}).
\]
Thus WLTS with prefix deletion also has linear search-space growth in reasoning depth under fixed search-budget parameters.
\end{corollary}

\begin{proof}
Each generated node is either kept or terminal; we bound the two contributions separately.
For kept nodes, fix any depth $t$.
Before the first deletion operation, the width invariant ensures that at most $W$ kept nodes can be generated at depth $t$.
Each deletion operation can free capacity in that layer, after which at most another $W$ kept nodes can be generated at the same depth before the next deletion.
Thus, after $C$ deletion operations, the total number of kept generated nodes at depth $t$ is at most $(C+1)W$.
Summing over all depths gives at most $(C+1)WD_{\max}$ kept nodes.
For terminal nodes, $|\mathcal{T}|\le N$ at termination, since the loop breaks once $|\mathcal{T}|\ge N$.
Summing the two contributions gives $G_{\mathrm{del}}\le (C+1)WD_{\max}+N$.
Finally, prefix deletion is triggered only after a terminal trajectory is collected, and the algorithm collects at most $N$ terminal trajectories; hence $C\le N$.
Substituting this bound gives the stated $O(NWD_{\max})$ result.
\end{proof}

\section{Implementation Details}
\subsection{Hyperparameters}

Unless otherwise specified, we use the default sampling parameters provided by each model.

For Qwen3-8B and Qwen3-14B, we set the temperature to 0.6 for the single-pass sampling baseline (no search) and to 0.7 for search algorithms to increase diversity.

For Llama-3.1-8B-Instruct, we use a temperature of 0 for the single-pass baseline (no search) and 0.6 for search algorithms to promote diversity.

The maximum answer length is set to 4096 tokens for Llama-3.1-8B-Instruct and 8192 tokens for Qwen3-8B and Qwen3-14B. For Llama-3.1-8B-Instruct, steps are segmented by line breaks. For Qwen3-8B and Qwen3-14B, we use a token-based segmentation scheme: each step contains 128 tokens for code-generation tasks and 512 tokens for other tasks, reflecting the difference in average response length across task types.

For Qwen3-8B and Qwen3-14B, we use non-thinking mode in the generator and thinking mode in the verifier, as an accurate verifier is required for verifier-guided search algorithms.

For the main comparison, tree-search baselines (MCTS, Q*, and ABMCTS-A) use a fixed selection budget $S=128$ and target up to $N=16$ terminal candidate trajectories.

All step-level methods share the proposer, verifier, step segmentation, and sampling parameters. They receive no separately trained task-specific value model or external feedback unavailable to the other methods.

\subsection{Benchmark Details}

For datasets with fewer than 100 problems, including AMC23, AIME24, and AIME25, we run each benchmark 10 times and report the average result. For all other datasets, we run multiple times to ensure the total sample size is at least 500, and then take the average. LiveCodeBench consists of code submissions collected from October 2024 to February 2025.
For ZebraLogic, we use the grid mode setting.

\subsection{Computation Cost}

This paper uses 80GB A100 GPUs for inference, with a cumulative inference time of approximately 7,500 hours across all experiments.

We use a multithreaded parallel approach to send multiple queries simultaneously. As a result, the total inference time on each dataset is approximately proportional to the total number of tokens processed. Specific wall-clock case experiments are shown in Table \ref{tab:runtime_metric}.

\section{Diversity Enhancement: Detailed Analysis}
\label{sec:appendix:diversity_details}

\textbf{Protocol.} Prefix deletion is intended to prevent WLTS from over-investing in a single high-scoring prefix family (Section~\ref{sec:wlts:diversity}). To isolate this mechanism, we sweep the deletion depth $h_{\mathrm{del}}\in\{1,2,4,8\}$ while fixing $(W,N,M)=(8,16,2)$, and evaluate Qwen3-8B and Qwen3-14B on AIME24, AIME25, and LiveCodeBench. We report both per-benchmark results and macro-averaged accuracy/token usage, since changing the deletion depth affects not only answer quality but also how quickly WLTS reaches repeated terminal answers.

\begin{wrapfigure}{r}{0.5\textwidth}
   \vspace{-\intextsep}
   \centering
   \includegraphics[width=0.48\textwidth]{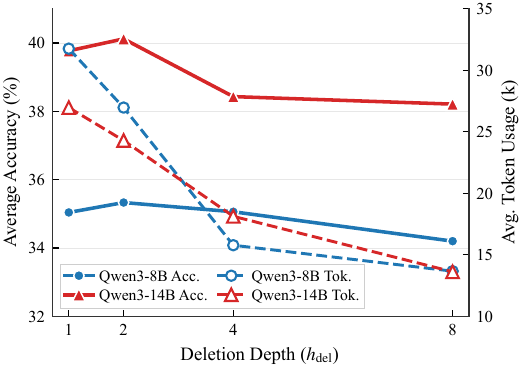}
   \caption{Effect of prefix deletion depth $h_{\mathrm{del}}$ on macro-averaged accuracy (solid lines, left axis) and token usage (dashed lines, right axis), evaluated on Qwen3-8B and Qwen3-14B over AIME24, AIME25, and LiveCodeBench.}
   \label{fig:diversity_ablation}
\end{wrapfigure}

\textbf{Results.} Figure~\ref{fig:diversity_ablation} and Table~\ref{tab:diversity_details} reveal two consistent patterns across both models. First, token usage decreases monotonically as $h_{\mathrm{del}}$ grows: deeper deletion truncates only narrow subtrees and leaves most of the explored prefix structure intact, so the freed capacity is spent extending nearby continuations rather than opening new high-level branches.

\newpage

Second, accuracy is non-monotonic in $h_{\mathrm{del}}$ and peaks at an intermediate value, confirming that prefix deletion functions as a diversity control rather than a pure pruning rule. Very shallow deletion erases broad prefix families together with their useful nearby continuations and forces exploration into distant regions that may not warrant the cost; very deep deletion preserves those continuations but exerts too little pressure to redirect search toward genuinely different high-level plans, so the algorithm keeps refining variants of an already-explored prefix. The intermediate setting balances these effects, retaining enough nearby continuations to capitalize on already-vetted prefix structure while still freeing capacity for substantially different alternatives---precisely the regime where prefix deletion is intended to operate. For deployment, choose $W$ according to the inference budget and start with $h_{\mathrm{del}}=1$--$2$, increasing it when token cost is the limiting factor.

\begin{table}[h]
   \caption{Prefix deletion depth ablation (Qwen3-8B and Qwen3-14B). Acc/Tok reports accuracy (\%) and average token usage (k).}
   \label{tab:diversity_details}
   \centering
   \small
   \renewcommand{\arraystretch}{1.15}
   \setlength{\tabcolsep}{4pt}
   \begin{tabular}{llcccc}
      \toprule
      Model & $h_{\mathrm{del}}$ & AIME24 & AIME25 & LiveCodeBench & Average \\
      \midrule
      \multirow{4}{*}{Qwen3-8B}
      & 1 & \underline{42.67}/42.91 & 29.33/38.21 & \textbf{33.13}/14.16 & 35.04/31.76 \\
      & 2 & \textbf{44.00}/34.35 & \underline{30.67}/31.42 & 31.33/15.15 & \textbf{35.33}/26.97 \\
      & 4 & 41.33/18.48 & \textbf{31.33}/17.19 & \underline{32.53}/11.67 & \underline{35.06}/15.78 \\
      & 8 & 40.00/15.23 & \underline{30.67}/15.72 & 31.93/10.05 & 34.20/13.67 \\
      \midrule
      \multirow{4}{*}{Qwen3-14B}
      & 1 & \textbf{46.00}/32.58 & \underline{35.33}/33.70 & \underline{37.99}/14.64 & \underline{39.77}/26.97 \\
      & 2 & 43.33/32.78 & \textbf{36.67}/29.26 & \textbf{40.36}/10.83 & \textbf{40.12}/24.29 \\
      & 4 & 40.67/22.02 & \textbf{36.67}/24.02 & 37.95/8.35 & 38.43/18.13 \\
      & 8 & \underline{44.67}/16.08 & 32.00/16.50 & 37.95/8.27 & 38.21/13.62 \\
      \bottomrule
   \end{tabular}
\end{table}

\section{Early Stopping: Detailed Analysis}
\label{sec:appendix:early_stopping}

\textbf{Protocol.} We ablate the agreement threshold $M$ in the early-stopping rule (Section~\ref{sec:wlts:earlystop}) on AIME24 and AIME25, where final answers can be canonicalized. For each setting, we report final accuracy, generated-token usage, the probability that early stopping is triggered, and the accuracy conditioned on early stopping. These metrics separate two effects: how often the search terminates before collecting all candidates, and how reliable the repeated answer is when termination occurs.

\begin{table}[h]
   \caption{Early stopping ablation on math benchmarks. Acc/Tok reports accuracy (\%) and average token usage (k). ES/ES Acc reports the early-stop rate (\%) and the accuracy conditioned on early stop (\%). Average is the arithmetic mean over AIME24 and AIME25.}
   \label{tab:early_stop_ablation}
   \centering
   \small
   \renewcommand{\arraystretch}{1.15}
   \setlength{\tabcolsep}{4pt}
   \begin{tabular}{llcccccc}
      \toprule
      Model & $M$
      & \multicolumn{2}{c}{AIME24}
      & \multicolumn{2}{c}{AIME25}
      & \multicolumn{2}{c}{Average} \\
      \cmidrule(lr){3-4}\cmidrule(lr){5-6}\cmidrule(lr){7-8}
      & & Acc/Tok & ES/ES Acc & Acc/Tok & ES/ES Acc & Acc/Tok & ES/ES Acc \\
      \midrule
      \multirow{3}{*}{Qwen3-8B}
      & 2 & 42.67/42.91 & 97.33/43.84 & \underline{29.33}/38.21 & 96.00/29.86 & 36.00/40.56 & 96.67/36.85\\
      & 3 & \underline{44.00}/68.26 & 84.67/\underline{50.39} & \underline{29.33}/54.13 & 80.67/\underline{35.54} & \underline{36.67}/61.20 & 82.67/\underline{42.97}\\
      & 4 & \textbf{44.33}/78.20 & 75.33/\textbf{50.44} & \textbf{30.00}/66.64 & 67.33/\textbf{41.58} & \textbf{37.17}/72.42 & 71.33/\textbf{46.01}\\
      \midrule
      \multirow{3}{*}{Qwen3-14B}
      & 2 & \textbf{46.00}/32.58 & 96.00/46.53 & 35.33/33.70 & 97.33/34.25 & \underline{40.67}/33.14 & 96.67/40.39\\
      & 3 & 44.00/60.95 & 88.67/\underline{49.62} & \underline{37.33}/58.50 & 84.00/\textbf{42.06} & \underline{40.67}/59.73 & 86.34/\underline{45.84}\\
      & 4 & \underline{45.33}/80.84 & 75.33/\textbf{54.87} & \textbf{37.67}/66.74 & 78.00/\underline{41.88} & \textbf{41.50}/73.79 & 76.67/\textbf{48.38}\\
      \bottomrule
   \end{tabular}%
\end{table}

\textbf{Results.} Table~\ref{tab:early_stop_ablation} exhibits two coupled effects of the agreement threshold $M$. First, raising $M$ shifts the early-stop rate downward and the conditional early-stop accuracy upward, confirming that repeated terminal answers become increasingly reliable as more independent trajectories agree on them, since a coincidental majority among incorrect answers is unlikely under sufficient agreement. Second, this added reliability is purchased at a disproportionate token cost: each additional required agreement compels WLTS to complete more candidates before consensus can form, and most of those extra trajectories are produced after the would-be majority answer has already emerged, so the search spends additional compute on confirmation rather than on discovering new answers. Consequently, the marginal final-accuracy gain from larger $M$ is small relative to the token overhead, and $M$ functions primarily as a deployment-time knob: small values are appropriate when budget pressure dominates, while larger values are justified only when a modest accuracy gain warrants substantially more inference compute.

\section{Inference-Time Scaling: Detailed Results for Qwen3-14B}
\label{sec:appendix:qwen14b_details}

\textbf{Protocol.} Table~\ref{tab:qwen14b_details} reports the detailed per-benchmark results underlying the inference-time scaling analysis in Section~\ref{sec:exp:scaling} (Figure~\ref{fig:inference_scaling}). We sweep compute for sequence search (BON and Beam Search), tree search (MCTS, Q*, and ABMCTS-A), and WLTS on Qwen3-14B. Acc/Tok reports accuracy (\%) and average generator token usage (k, in thousands); all runs use temperature $t{=}0.7$ and maximum generation length $l{=}8192$.

\begin{table}[h]
   \caption{Inference-time scaling ablation on Qwen3-14B: detailed results for sequence search (BON/beam), tree search (MCTS/Q*/ABMCTS-A), and WLTS under different compute settings. The value before the slash is the accuracy, and the value after the slash is the token usage (k).}
   \label{tab:qwen14b_details}
   \centering
   \small
   \renewcommand{\arraystretch}{1.15}
   \setlength{\tabcolsep}{3pt}
   \begin{tabular}{llccccc}
      \toprule
      Method & Compute setting & AIME24 & AIME25 & LiveCodeBench & ZebraLogic & Average \\
      \midrule
      \multirow{4}{*}{BON}
      & $N{=}2$ & 35.33/5.58 & 30.00/5.21 & 30.72/0.97 & 48.10/4.28 & 36.04/4.01 \\
      & $N{=}4$ & 37.33/12.63 & 28.67/11.22 & 33.13/2.03 & 49.90/8.57 & 37.26/8.61 \\
      & $N{=}8$ & 42.00/24.98 & 30.67/22.02 & 30.72/4.65 & 49.70/17.09 & 38.27/17.19 \\
      & $N{=}16$ & 40.67/56.16 & 32.00/46.06 & 31.93/9.09 & 50.60/34.22 & 38.80/36.38 \\
      \midrule
      \multirow{4}{*}{Beam}
      & $W{=}2$, $b{=}1$ & 34.00/6.06 & 28.67/5.50 & 30.72/1.18 & 49.80/4.27 & 35.80/4.25 \\
      & $W{=}2$, $b{=}2$ & 38.00/14.47 & \underline{34.00}/12.93 & 33.73/3.44 & 52.40/8.86 & 39.53/9.93 \\
      & $W{=}4$, $b{=}2$ & 42.67/30.60 & 32.00/27.92 & 34.94/6.13 & 55.50/18.38 & 41.28/20.76 \\
      & $W{=}4$, $b{=}4$ & 40.00/76.74 & 33.33/62.52 & 34.34/18.55 & 55.10/33.84 & 40.70/47.91 \\
      \midrule
      \multirow{3}{*}{MCTS}
      & $S{=}32$, $N{=}4$ & 40.67/40.46 & 26.67/41.99 & 33.73/8.29 & 53.80/40.81 & 38.72/32.89 \\
      & $S{=}64$, $N{=}8$ & 41.33/86.21 & 32.67/82.58 & 33.73/22.65 & \underline{56.20}/68.71 & 40.98/65.04 \\
      & $S{=}128$, $N{=}16$ & 42.00/155.00 & 33.33/152.75 & 34.94/40.98 & 54.40/112.89 & 41.17/115.41 \\
      \midrule
      \multirow{3}{*}{Q*}
      & $S{=}32$, $N{=}4$ & 36.66/14.72 & 30.67/13.77 & 33.13/9.38 & 49.80/17.50 & 37.57/13.84 \\
      & $S{=}64$, $N{=}8$ & 38.33/31.54 & 31.33/28.45 & 36.75/21.25 & 50.70/20.98 & 39.28/25.55 \\
      & $S{=}128$, $N{=}16$ & 38.00/57.28 & 33.66/51.57 & \textbf{38.55}/43.42 & 51.40/62.38 & 40.40/53.66 \\
      \midrule
      \multirow{3}{*}{ABMCTS-A}
      & $S{=}64$, $N{=}4$ & 37.33/27.81 & 29.33/28.71 & 31.33/19.66 & 49.60/25.78 & 36.90/25.49 \\
      & $S{=}128$, $N{=}8$ & 38.00/58.30 & 29.33/57.79 & 34.94/41.22 & 52.50/51.09 & 38.69/52.10 \\
      & $S{=}256$, $N{=}16$ & 42.67/117.78 & 32.00/114.20 & 37.35/83.27 & 51.00/102.26 & 40.76/104.38 \\
      \midrule
      \multirow{3}{*}{WLTS}
      & $W{=}2$, $N{=}2$ & 38.67/8.04 & 31.33/7.47 & 30.12/1.11 & 53.20/5.73 & 38.33/5.59 \\
      & $W{=}4$, $N{=}4$ & \underline{43.33}/17.42 & \underline{34.00}/16.62 & 37.35/2.10 & 55.10/14.27 & \underline{42.45}/12.60 \\
      & $W{=}8$, $N{=}16$ & \textbf{46.00}/32.58 & \textbf{35.33}/33.70 & \underline{37.99}/14.64 & \textbf{63.60}/37.69 & \textbf{45.73}/29.65 \\
      \bottomrule
   \end{tabular}
\end{table}

\textbf{Results.} Table~\ref{tab:qwen14b_details} reveals distinct scaling behaviors across method families, all of which trace back to how each algorithm uses additional budget. Best-of-N samples whole trajectories, so additional budget broadens the candidate pool but cannot revise erroneous prefixes shared across samples; its curve flattens once the dominant failure modes saturate the candidate set. Beam Search benefits from step-level scoring and improves sharply at moderate widths, but at larger widths its irreversible pruning becomes the binding constraint: extra capacity is consumed by minor variants of the surviving frontier and cannot recover from prematurely committed prefixes, so accuracy stalls or even regresses despite the larger budget. Standard tree search restores backtracking but spends most of its added budget on shallow nodes, since unconstrained branching leaves deeper layers visit-starved; MCTS therefore consumes substantially more tokens for marginal gains, and Q* and ABMCTS-A trim some of this cost but inherit the same shallow bias and remain below WLTS at comparable budgets. WLTS exhibits the steepest and most sustained improvement because the per-layer width cap forces each additional expansion to deepen an existing path or open a new prefix family rather than thicken an already-saturated layer, so extra compute translates directly into deeper search and more high-quality completed candidates while backtracking preserves the recovery options that sequence search lacks. The advantage is broad rather than benchmark-specific: WLTS leads on the majority of benchmarks at every budget level, with the only crossover occurring on LiveCodeBench at high budget, where Q* attains a marginally higher score using several times more tokens.

\section{Token-Budget-Matched Comparison}
\label{sec:appendix:token_matched}

\textbf{Protocol.} The main comparison in Table~\ref{tab:combined_results} matches the target number of terminal candidates, but realized token usage differs because each search method allocates budget differently. WLTS may stop early on canonicalizable tasks, whereas tree-search baselines can spend several times more tokens before producing the same number of terminal candidates. To complement the candidate-count-matched protocol, we additionally compare methods within a similar token range on Qwen3-14B.

We add a Best-of-N run with $N{=}10$ so that its average token usage falls near the Core/Full WLTS range. For the remaining baselines, we select the configurations from Table~\ref{tab:qwen14b_details} with average token usage closest to this range: Beam Search with $W{=}4,b{=}2$, MCTS with $S{=}32,N{=}4$, Q* with $S{=}64,N{=}8$, and ABMCTS-A with $S{=}64,N{=}4$, where $S$ denotes the number of tree-search selection iterations. We also include Core WLTS, which disables both prefix deletion and agreement-based early stopping, to isolate the effect of the width-limited backtracking structure.

\begin{table}[h]
   \caption{Token-budget-matched comparison on Qwen3-14B. Each cell reports accuracy (\%) / generated tokens (k) per problem. BON uses an additional $N{=}10$ run, and the other baselines are selected from Table~\ref{tab:qwen14b_details} to have token usage near the Core/Full WLTS range. ``Core WLTS'' disables both prefix deletion and agreement-based early stopping.}
   \label{tab:matched_budget}
   \centering
   \small
   \setlength{\tabcolsep}{4pt}
   \begin{tabular}{lccccc}
      \toprule
      Method & AIME24 & AIME25 & LiveCodeBench & ZebraLogic & Average \\
      \midrule
      Best-of-N $N{=}10$           & 40.67 / 31.96 & 31.33 / 27.93 & 35.54 / 7.14  & 50.00 / 21.49 & 39.39 / 22.13 \\
      Beam Search $W{=}4, b{=}2$   & \underline{42.67} / 30.60 & 32.00 / 27.92 & 34.94 / 6.13  & 55.50 / 18.38 & 41.28 / 20.76 \\
      MCTS $S{=}32, N{=}4$         & 40.67 / 40.46 & 26.67 / 41.99 & 33.73 / 8.29  & 53.80 / 40.81 & 38.72 / 32.89 \\
      Q* $S{=}64, N{=}8$           & 38.33 / 31.54 & 31.33 / 28.45 & 36.75 / 21.25 & 50.70 / 20.98 & 39.28 / 25.55 \\
      ABMCTS-A $S{=}64, N{=}4$     & 37.33 / 27.81 & 29.33 / 28.71 & 31.33 / 19.66 & 49.60 / 25.78 & 36.90 / 25.49 \\
      \midrule
      \wltsrow Core WLTS         & 41.33 / 27.28 & \underline{32.33} / 25.88 & \underline{37.35} / 4.09  & \underline{55.80} / 26.51 & \underline{41.70} / 20.94 \\
      \wltsrow \textbf{WLTS}           & \textbf{46.00} / 32.58 & \textbf{35.33} / 33.70 & \textbf{37.99} / 14.64 & \textbf{63.60} / 37.69 & \textbf{45.73} / 29.65 \\
      \bottomrule
   \end{tabular}%
\end{table}

\textbf{Results.} Table~\ref{tab:matched_budget} confirms that the WLTS advantage is not an artifact of the candidate-count-matched protocol. Core WLTS already matches or exceeds the strongest non-WLTS baseline at essentially the same average token budget, isolating the per-layer width cap and backtracking structure as the primary source of the accuracy--efficiency gain rather than the optional deployment components. The underlying mechanism is the same one identified in the main text: a strict per-layer cap forces extra expansions to deepen the search rather than thicken shallow layers, so under any fixed token budget a larger fraction of compute lands on deeper steps where verifier supervision is otherwise weakest, while backtracking allows that deeper compute to revise rather than only extend earlier prefixes. Adding prefix deletion and agreement-based early stopping shifts Full WLTS into a slightly higher token regime but yields a substantially larger accuracy gain, indicating that the optional components broaden the prefix space and trim redundant exploration without altering the core advantage. The fact that WLTS leads on every benchmark in this table further suggests that the gain is structural rather than benchmark-specific, mirroring the picture from the candidate-count-matched protocol.

\section{Runtime Metric}
\label{sec:appendix:runtime}

The main results in Table~\ref{tab:combined_results} report computational cost via generator token usage, which gives a hardware-agnostic view but is not equivalent to wall-clock latency. Latency also depends on the serving stack and the parallelism structure of each search method: in modern LLM serving, search-based methods can still benefit from concurrent requests and prefix KV-cache reuse, but their parallelism is generally less direct than Best-of-N due to stronger inter-step dependencies. Best-of-N is embarrassingly parallel across the $N$ trajectories; sequence search such as Beam Search exposes width-level parallelism per step but is sequential across steps; and backtracking-based tree search (MCTS, Q*, WLTS) introduces additional inter-step dependencies that limit how much computation can be overlapped in time.

\textbf{Setup.} To complement the token-usage view in Table~\ref{tab:combined_results}, we additionally measure wall-clock runtime on Llama-3.1-8B (Skywork-PRM verifier) under the same multithreaded $4{\times}$A800 setup used for the main experiments, with all other settings unchanged. Table~\ref{tab:runtime_metric} reports per-benchmark accuracy, generator tokens, and wall-clock time per problem on MATH500, OlympiadBench, AMC23, and MBPP, together with their macro-averages.

\begin{table}[h]
   \caption{Per-benchmark wall-clock runtime on Llama-3.1-8B (Classifier RM setting) under a multithreaded $4{\times}$A800 setup. Acc.\ is accuracy (\%); Tok.\ is the average generated tokens per problem (k); Time is the average wall-clock completion time per problem (seconds). The Average column reports macro-averaged values across the four benchmarks; parentheses next to accuracy indicate absolute improvements over Best-of-N (in percentage points), and parentheses next to Tok.\ and Time indicate cost relative to Best-of-N (= 100\%).}
   \label{tab:runtime_metric}
   \centering
   \small
   \renewcommand{\arraystretch}{1.15}
   \setlength{\tabcolsep}{3pt}
   \begin{tabular}{llccccc}
      \toprule
      Method & Metric & MATH500 & Olympiad Bench & AMC23 & MBPP & Average \\
      \midrule
      \multirow{3}{*}{Best-of-N}
      & Acc.(\%) & 66.20 & 28.30 & 50.00 & 72.50 & 54.25 \\
      & Tok.(k)  & 12.68 & 23.16 & 20.24 &  4.42 & 15.13 (100\%) \\
      & Time(s)  &  1.06 &  1.76 &  1.62 &  0.32 &  1.19 (100\%) \\
      \midrule
      \multirow{3}{*}{Beam Search}
      & Acc.(\%) & 70.20 & 33.33 & 52.50 & 71.70 & 56.93 (+2.68) \\
      & Tok.(k)  &  9.87 & 14.47 & 13.08 &  4.10 & 10.38 (69\%) \\
      & Time(s)  &  2.76 &  3.84 &  3.31 &  0.73 &  2.66 (224\%) \\
      \midrule
      \multirow{3}{*}{MCTS}
      & Acc.(\%) & 66.40 & 30.67 & 55.00 & 72.50 & 56.14 (+1.89) \\
      & Tok.(k)  & 68.93 & 86.37 & 91.65 & 13.39 & 65.09 (430\%) \\
      & Time(s)  & 16.21 & 20.86 & 21.77 &  3.46 & 15.58 (1309\%) \\
      \midrule
      \multirow{3}{*}{Q*}
      & Acc.(\%) & 69.80 & 32.74 & 55.00 & 73.10 & 57.66 (+3.41) \\
      & Tok.(k)  & 57.15 & 60.39 & 62.46 & 14.22 & 48.56 (321\%) \\
      & Time(s)  & 14.23 & 14.43 & 15.34 & 12.40 & 14.10 (1185\%) \\
      \midrule
      \multirow{3}{*}{ABMCTS-A}
      & Acc.(\%) & 66.20 & 30.39 & 52.50 & 70.90 & 55.00 (+0.75) \\
      & Tok.(k)  & 38.16 & 42.48 & 41.34 & 18.39 & 35.09 (232\%) \\
      & Time(s)  & 15.71 & 17.39 & 17.23 &  7.47 & 14.45 (1214\%) \\
      \midrule
      \wltsrow & Acc.(\%) & \textbf{72.40} & \textbf{36.30} & \textbf{65.00} & \textbf{75.90} & \textbf{62.40} (+8.15) \\
      \wltsrow & Tok.(k)  & 16.98 & 30.54 & 30.05 &  7.57 & 21.29 (141\%) \\
      \wltsrow \multirow{-3}{*}{\textbf{WLTS}} & Time(s) &  \textbf{6.81} & \textbf{11.76} & \textbf{11.43} &  3.17 & \textbf{8.29} (697\%) \\
      \bottomrule
   \end{tabular}
\end{table}

\textbf{Results.} Table~\ref{tab:runtime_metric} matches the qualitative picture from the token-cost view but exposes two effects that token counts alone cannot reveal. Among backtracking-based methods, WLTS achieves the best accuracy--latency trade-off because the per-layer width cap upper-bounds the work performed at each depth and therefore the number of sequential expansion rounds needed to fill the search tree, whereas unconstrained tree search must service many more shallow expansions before producing comparable terminal candidates. Best-of-N remains the fastest in raw latency due to its embarrassingly parallel structure across complete trajectories, and Beam Search retains a latency edge through per-step batching of frontier candidates; both, however, omit backtracking entirely and pay for that absence in final accuracy. The takeaway is therefore not that WLTS is universally faster than non-backtracking baselines, but that the per-layer width cap turns backtracking from a latency liability into a controllable cost, making WLTS the most practical option whenever recovery from erroneous prefixes is required. The three unconstrained tree-search baselines (MCTS, Q*, and ABMCTS-A) all fall in a similar 14--16\,s wall-clock band per problem despite their different expansion heuristics, indicating that their shared inter-step dependency structure---rather than per-method micro-optimizations---is what dominates runtime; WLTS sits below this band at 8.29\,s while still delivering the highest accuracy.

Each generated step-level node triggers a proposer call and a verifier call, with decoding bounded by the proposer's per-step token budget and the generative verifier's response cap. Longer prefixes increase prefill computation, while prefix caching allows shared prefixes to reuse their keys and values. Profiling WLTS under its main configuration on Qwen3-8B and AIME25 shows that total prefill tokens are 3.67 times total generated tokens, yet the prefill-to-decode time ratio is only 0.0065. Decoding therefore dominates the measured runtime, supporting generated-node count as a practical compute proxy in this setting. This correspondence weakens when decoding no longer dominates, such as with very short generations and disproportionately long prompts, or when prefix-cache reuse differs sharply across methods.

\section{Comparison with REBASE and ABMCTS-M}

REBASE~\citep{REBASE2025} is a reward-guided extension of beam search that allocates the remaining sampling budget among prefixes at each depth according to softmax-normalized rewards. We evaluate it on both Qwen3 models with a layer-level expansion budget of 16. Table~\ref{tab:rebase_comparison} compares it with Beam Search and WLTS. The results show that WLTS obtains higher mean accuracy with fewer generator tokens than REBASE on both models.

\begin{table}[ht]
\centering\small
\setlength{\tabcolsep}{3pt}
\caption{Comparison with REBASE. Cells report accuracy (\%) / generator tokens (k).}
\label{tab:rebase_comparison}
\begin{tabular}{llccccc}
\toprule
Model & Method & AIME24 & AIME25 & ZebraLogic & LiveCodeBench & Mean \\
\midrule
Qwen3-8B & Beam Search & 32.67/72.8 & 27.33/55.2 & 45.80/52.6 & 31.93/16.5 & 34.43/49.26 \\
& REBASE & 35.33/57.0 & 28.67/50.6 & 44.00/46.8 & 30.12/13.7 & 34.53/42.03 \\
& WLTS & 42.67/42.9 & 29.33/38.2 & 48.40/36.4 & 33.13/14.2 & 38.38/32.92 \\
\midrule
Qwen3-14B & Beam Search & 40.00/76.7 & 33.33/62.5 & 55.10/33.8 & 34.34/18.6 & 40.70/47.91 \\
& REBASE & 42.67/55.5 & 32.67/50.0 & 52.60/36.7 & 34.94/13.3 & 40.72/38.88 \\
& WLTS & 46.00/32.6 & 35.33/33.7 & 63.60/37.7 & 37.99/14.6 & 45.73/29.65 \\
\bottomrule
\end{tabular}
\end{table}

We additionally evaluate another ABMCTS variant, ABMCTS-M~\citep{WiderDeeper2025}, with Qwen3-8B under the shared-verifier step-level interface. The two variants trade wins across tasks and have similar macro-averages, while WLTS has the highest macro-average (Table~\ref{tab:abmcts_variants}).

\begin{table}[ht]
\centering\small
\caption{Accuracy (\%) of ABMCTS variants and WLTS on Qwen3-8B.}
\label{tab:abmcts_variants}
\begin{tabular}{lccccc}
\toprule
Method & AIME24 & AIME25 & LiveCodeBench & ZebraLogic & Mean \\
\midrule
ABMCTS-M & 34.00 & 26.67 & 27.71 & 44.80 & 33.30 \\
ABMCTS-A & 32.67 & 29.33 & 29.52 & 40.80 & 33.08 \\
WLTS & 42.67 & 29.33 & 33.13 & 48.40 & 38.38 \\
\bottomrule
\end{tabular}
\end{table}

\section{Additional Budget Tiers}

Table~\ref{tab:additional_budget_tiers} reports four operating tiers on AIME24 with Qwen3-14B. The tiers group runs with broadly comparable realized generator-token costs. WLTS has the highest observed accuracy in each tier.

\begin{table}[ht]
\centering\small
\caption{Multi-budget comparison. Cells report accuracy (\%) / generator tokens (k).}
\label{tab:additional_budget_tiers}
\begin{tabular}{lcccc}
\toprule
Method & Small & Medium & Large & Extra Large \\
\midrule
Best-of-N & 36.33/9.01 & 37.66/19.37 & 37.33/38.43 & 42.67/75.96 \\
Beam Search & 38.00/14.47 & 38.67/22.97 & 42.67/30.60 & 40.00/76.74 \\
MCTS & 36.67/10.88 & 37.66/21.03 & 40.67/40.46 & 41.33/86.21 \\
Q* & 35.33/10.52 & 34.33/18.83 & 38.33/31.54 & 39.33/71.29 \\
WLTS & 38.67/8.04 & 43.33/17.42 & 46.00/32.58 & 48.67/67.53 \\
\bottomrule
\end{tabular}
\end{table}

\section{Maximum Verifier-Supervised Depth}
\label{sec:appendix:depth_diagnostic}

We record the deepest verifier-supervised node on each of 1,000 ZebraLogic problems under the main-table Qwen3-14B setting. Table~\ref{tab:supervised_depth} reports the mean maximum depth for small, medium, and large grids. The methods reach similar depths on small grids, while WLTS reaches deeper supervised states on medium and large grids.

\begin{table}[ht]
\centering\small
\caption{Mean maximum verifier-supervised depth (steps) on ZebraLogic, grouped by grid size.}
\label{tab:supervised_depth}
\begin{tabular}{lccc}
\toprule
Method & Small & Medium & Large \\
\midrule
ABMCTS-A & 2.1 & 2.6 & 2.8 \\
MCTS & 2.3 & 3.9 & 4.7 \\
Q* & 2.5 & 4.9 & 7.6 \\
WLTS & 2.4 & 5.7 & 10.4 \\
\bottomrule
\end{tabular}
\end{table}

\section{Step Segmentation}

We vary only step length on Qwen3-14B for AIME24 and AIME25 (Table~\ref{tab:step_sensitivity}). Finer segmentation improves the reported mean accuracy while increasing generator-token usage. The 512-token setting provides a lower-cost operating point among these configurations.

\begin{table}[ht]
\centering\small
\caption{Step-length sensitivity on Qwen3-14B.}
\label{tab:step_sensitivity}
\begin{tabular}{rcccc}
\toprule
Step length (tokens) & AIME24 (\%) & AIME25 (\%) & Mean Acc. (\%) & Mean Tok. (k) \\
\midrule
128 & 48.67 & 38.00 & 43.34 & 72.4 \\
256 & 47.33 & 36.00 & 41.67 & 48.0 \\
512 & 46.00 & 35.33 & 40.67 & 33.1 \\
\bottomrule
\end{tabular}
\end{table}

\section{Selection Priority}
\label{sec:appendix:selection_sensitivity}

All main WLTS experiments use $c_{\mathrm{uct}}=0$, giving value-only selection with ties broken toward deeper nodes. Table~\ref{tab:selection_sensitivity} compares priorities on Qwen3-14B. Overall, all three settings perform well.

\begin{table}[!ht]
\centering\small
\caption{Selection-rule sensitivity on Qwen3-14B.}
\label{tab:selection_sensitivity}
\begin{tabular}{lccc}
\toprule
Selection rule & AIME24 (\%) & AIME25 (\%) & Mean Tok. (k) \\
\midrule
Value only, $c_{\mathrm{uct}}=0$ & 46.00 & 35.33 & 33.1 \\
Global bonus, $c_{\mathrm{uct}}=0.5$ & 44.67 & 34.00 & 35.7 \\
Global bonus, $c_{\mathrm{uct}}=1.0$ & 45.00 & 35.67 & 31.9 \\
\bottomrule
\end{tabular}
\end{table}

\clearpage
\section{Generative Verifier Prompts}

\begin{tcolorbox}[title=Prompt Template for Generative Verifier,fonttitle=\bfseries,breakable,enhanced,
boxrule=0.8mm,
segmentation style={solid,draw=black,line width=0.8mm},
colback=white,
colframe=black,
colbacktitle=white,
coltitle=black,
pad at break=0mm]
\begin{Verbatim}[
  fontsize=\small,
  breaklines,
  commandchars=\\\{\},
  breakindent=0pt,
  breaksymbolleft={},
  breaksymbolright={}
]
[User]
{\color{blue}{\{QUESTION\}}}

[Assistant]
{\color{blue}{\{CANDIDATE\_REASONING\_OR\_SOLUTION\}}}

[User]
You are a careful and precise verifier. Your task is to assess whether the assistant's reasoning so far is logically and mathematically correct. You must not attempt to solve the problem yourself or continue the reasoning. Only check whether the existing reasoning is correct step by step. At the end of your response, on a new line, write exactly:
"Is the existing reasoning correct (Yes/No)? X"
where X is either Yes or No. Only write Yes if all steps so far are logically correct.
\end{Verbatim}
\end{tcolorbox}

\end{document}